\documentclass{article}

\usepackage{microtype}
\usepackage{graphicx}
\usepackage{subfigure}
\usepackage{booktabs} 

\usepackage{hyperref}
\usepackage{url}
\usepackage{color, colortbl}
\usepackage{amssymb}
\usepackage{adjustbox}
\usepackage{amsthm}
\usepackage{bm}
\usepackage{tikz}
\usepackage{xcolor}
\usepackage{multirow}
\definecolor{LightCyan}{rgb}{0.88,1,1}
\usepackage{amsmath}
\usepackage[misc]{ifsym}

\usepackage[accepted]{icml2025}

\usepackage{amsmath}
\usepackage{amssymb}
\usepackage{mathtools}
\usepackage{amsthm}

\usepackage[capitalize,noabbrev]{cleveref}

\theoremstyle{plain}
\newtheorem{theorem}{Theorem}[section]

\newtheorem{lemma}[theorem]{Lemma}

\theoremstyle{definition}
\newtheorem{definition}[theorem]{Definition}

\theoremstyle{remark}

\usepackage[textsize=tiny]{todonotes}

\icmltitlerunning{Context Matters: Query-aware Dynamic Long Sequence Modeling of Gigapixel Images}

\begin{document}

\twocolumn[
\icmltitle{Context Matters: \\ Query-aware Dynamic Long Sequence Modeling of Gigapixel Images}




\begin{icmlauthorlist}
\icmlauthor{Zhengrui Guo}{ust,bici}
\icmlauthor{Qichen Sun}{pku}
\icmlauthor{Jiabo Ma}{ust}
\icmlauthor{Lishuang Feng}{bici}
\icmlauthor{Jinzhuo Wang}{pku}
\icmlauthor{Hao Chen}{ust}
\end{icmlauthorlist}

\icmlaffiliation{ust}{Hong Kong University of Science and Technology, HK, China}
\icmlaffiliation{bici}{Beijing Institute of Collaborative Innovation, Beijing, China}
\icmlaffiliation{pku}{Peking University, Beijing, China}

\icmlcorrespondingauthor{Zhengrui Guo}{zguobc@connect.ust.hk}
\icmlcorrespondingauthor{Hao Chen}{jhc@cse.ust.hk}

\icmlkeywords{Computational Pathology, Multi-instance Learning, Whole Slide Images}

\vskip 0.3in
]



\printAffiliationsAndNotice{\icmlEqualContribution} 

\begin{abstract}
Whole slide image (WSI) analysis presents significant computational challenges due to the massive number of patches in gigapixel images. While transformer architectures excel at modeling long-range correlations through self-attention, their quadratic computational complexity makes them impractical for computational pathology applications. Existing solutions like local-global or linear self-attention reduce computational costs but compromise the strong modeling capabilities of full self-attention. 
In this work, we propose \textbf{Querent}, \textit{i.e.}, the \textbf{quer}y-awar\textbf{e} long co\textbf{nt}extual dynamic modeling framework, which achieves a theoretically bounded approximation of full self-attention while delivering practical efficiency.
Our method adaptively predicts which surrounding regions are most relevant for each patch, enabling focused yet unrestricted attention computation only with potentially important contexts. By using efficient region-wise metadata computation and importance estimation, our approach dramatically reduces computational overhead while preserving global perception to model fine-grained patch correlations. Through comprehensive experiments on biomarker prediction, gene mutation prediction, cancer subtyping, and survival analysis across over 10 WSI datasets, our method demonstrates superior performance compared to the state-of-the-art approaches. Codes are 
\href{https://github.com/dddavid4real/Querent}{here}.
\end{abstract}

\section{Introduction}
Computational pathology (CPath) represents a transformative shift in clinical diagnostics, leveraging artificial intelligence and deep learning to analyze the growing collections of whole slide images (WSIs) from medical facilities \citep{van2021deep,cui2021artificial,zheng2025diffusion}. By digitalizing traditional pathology workflows, this emerging field enhances clinical decision-making through more standardized diagnoses, enables the identification of new biomarkers, and helps predict treatment outcomes for patients \citep{niazi2019digital,song2023artificial}. These WSIs, also known as gigapixel images, typically contain between $10,000^2\sim100,000^2$ pixels. The challenge lies in identifying critical diagnostic features that may be dispersed across various tissue regions within these highly informative images, \textit{i.e.}, analogous to finding a needle in a haystack \citep{jin2023label}. These unique challenges have driven the development of multi-instance learning (MIL), a weakly-supervised learning paradigm for WSI analysis \citep{campanella2019clinical,lu2021data,shao2021transmil,xu2023multimodal,zhou2023cross,yang2024mambamil}.

\begin{figure}
    \centering
    \includegraphics[width=1\linewidth]{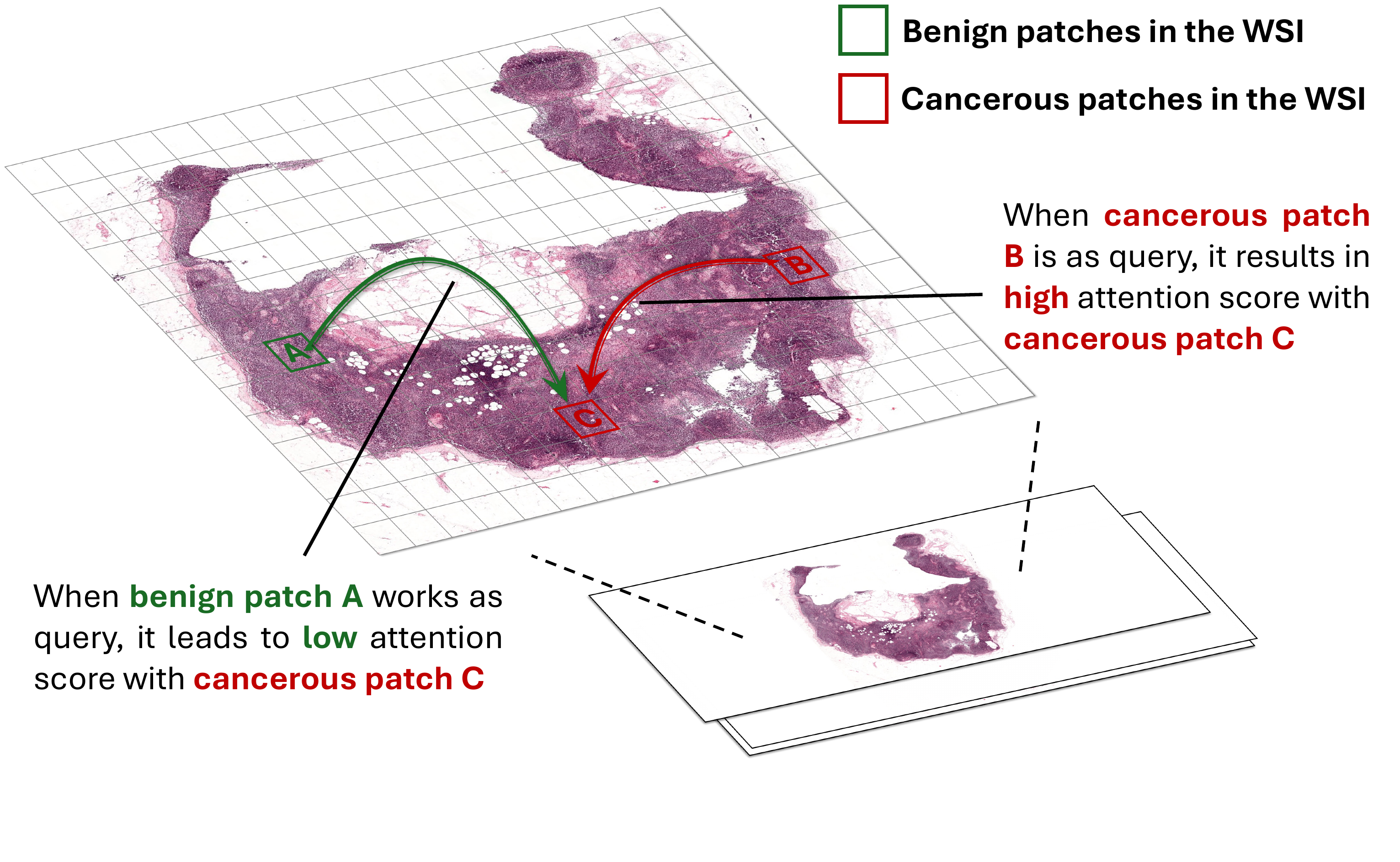}
    \vspace{-1.2cm}
    \caption{Illustration of context-dependent patch relationships in whole slide images. When a benign patch (A) interacts with cancerous patch C, it shows low correlation, while a cancerous patch (B) shows high correlation with patch C. This demonstrates how the same patch (C) can have fundamentally different relationships with other patches depending on the biological context.}
    \label{fig:1}
    \vspace{-0.4cm}
\end{figure}

MIL performs slide-level analysis through three key steps: (1) segmenting and cropping tissues in a WSI into smaller image patches; (2) extracting representation features from these patches using a pretrained encoder (typically a CPath foundation model \citep{chen2024towards,lu2024visual,ma2024towards,xu2024multimodal,du2025beyond,xiong2025survey}); and (3) aggregating the patch features through specific principles to obtain the slide-level representation. Among various feature aggregation approaches, transformer architectures have emerged as a particularly promising solution due to their powerful ability to model long-range dependencies through self-attention mechanisms \citep{vaswani2017attention,dosovitskiy2020image}. This capability is especially valuable in WSI analysis, where diagnostically relevant features often manifest through complex spatial relationships between distant tissue regions. However, the quadratic computational complexity $\mathbf{O}(n^2)$ of the standard transformer's self-attention mechanism presents a significant challenge when applied to WSIs, as a typical slide can contain thousands to tens of thousands of patches \citep{wang2023image,song2024multimodal,guo2024focus,xiong2025beyond}.

To address this computational barrier, several adaptations of transformer architectures have been proposed for WSI analysis. These include linear attention methods that reduce complexity to $\mathbf{O}(n)$ \citep{shao2021transmil}, and local-global attention approaches that restrict attention computation to predetermined spatial patterns \citep{chen2022scaling,li2024longmil}. While these modifications successfully reduce computational overhead, they inevitably compromise the transformer's inherent modeling capabilities. As demonstrated by \citet{dao2022flashattention} and \citet{han2023flatten}, linear approximation of self-attention leads to only sub-optimal modeling performance. Meanwhile, local-global attention makes strong assumptions about which spatial relationships are important, failing to adapt to the highly variable and context-dependent nature of pathological features in WSIs.

These limitations motivate the need for a more adaptive approach to modeling inter-patch relationships in WSI analysis. A key observation is that the relevance of surrounding tissue regions varies significantly depending on which specific region is being examined \citep{heindl2015mapping,yuan2016spatial}. As shown in Fig. \ref{fig:1}, when analyzing a tumor boundary region, nearby regions showing the tumor-stroma interface might be highly relevant, while distant regions of normal tissue might be less informative. Conversely, when examining an area of inflammation, regions with similar inflammatory patterns across the slide might be more relevant than adjacent but histologically different regions. This context-dependent nature of patch relationships suggests that an ideal attention mechanism should dynamically prioritize relevant interactions for each query patch while maintaining the capability to model long-range dependencies when necessary. 

Based on this insight, we propose \textbf{Querent}, a novel framework for dynamic long-range contextual modeling of gigapixel images through adaptive determination of patch relationships. Our approach maintains the modeling power of full attention while achieving computational efficiency through dynamic sparsification. Rather than using fixed patterns or uniform approximations, it estimates the potential importance of patch relationships through efficient region metadata computation and selectively applies full attention to the most relevant interactions. This query-aware strategy allows each patch to have its unique attention pattern, better capturing the heterogeneous nature of histological features while remaining computationally tractable for gigapixel WSIs. Theoretically, we prove that Querent's query-aware attention mechanism maintains expressiveness within a small constant bound of full self-attention. Empirically, we demonstrate Querent's effectiveness across multiple CPath tasks, including biomarker prediction, gene mutation prediction, cancer subtyping, and survival prediction, where it consistently outperforms state-of-the-art models on over 10 WSI datasets. 

The main contributions of this work are as follows: (1) We propose a novel query-aware attention mechanism that dynamically adapts to each patch's unique context in gigapixel WSIs, maintaining full attention's expressiveness while achieving computational efficiency; (2) We develop efficient region-level metadata summarization and importance estimation modules that enables dynamic sparsification of attention patterns while preserving modeling capabilities; (3) We establish Querent's effectiveness through theoretical bounds on its expressiveness and extensive empirical validation across diverse CPath tasks.

\section{Related Work}
Given the massive image size of WSIs and GPU memory restrictions \citep{araujo2019computing}, researchers have widely adopted MIL for WSI analysis. Recent MIL-based approaches have achieved promising results in diagnosing diseases and predicting patient outcomes under the formulation of weakly-supervised learning \citep{campanella2019clinical,chikontwe2020multiple,li2021dt,xiang2022dsnet,hou2022h,zheng2022graph,wang2022scl,yu2023prototypical,xiong2023diagnose,lin2023interventional,xiong2024mome,song2024morphological,xiong2024takt}. 

MIL primarily addresses the challenge of aggregating patch-level features into a comprehensive slide-level representation for diagnostic purposes. Early MIL approaches employed straightforward, non-parametric aggregation techniques, such as max and mean pooling operations, to combine these features \citep{campanella2019clinical}. Recent advances in MIL have emphasized the development of more sophisticated aggregation strategies to effectively capture diagnostic patterns scattered across numerous patches. ABMIL \citep{ilse2018attention} introduces an attention-based framework that computes importance weights for each patch using a side network during the aggregation process. CLAM \citep{lu2021data} enhanced this attention mechanism by incorporating clustering-constrained learning and flexible attention branches to enhance WSI classification performance. DSMIL \citep{li2021dual} develops a dual-stream architecture that capitalizes on the hierarchical structure of WSIs by integrating features across multiple magnification levels. Further, DTFD-MIL \citep{zhang2022dtfd} introduces a novel double-tier framework that leverages pseudo-bags to maximize feature utilization and minimize the bag-instance imbalance problem in WSI analysis. Taking a graph-based perspective, WiKG \citep{li2024dynamic} reformulates WSI analysis by representing patches as nodes in a knowledge graph and utilizing head-to-tail embeddings to generate dynamic graph representations. Combined with the State Space Models \citep{gu2023mamba}, MambaMIL \citep{yang2024mambamil} features a Sequence Reordering Mamba module that processes instance sequences through both original and reordered pathways to enhance long-range dependency modeling while maintaining linear complexity. The field has further evolved with the integration of multimodal information, as researchers have begun incorporating pathology image captions \citep{lu2023visual}, diagnostic reports \citep{guo2024histgen}, as well as genomic profiles \citep{sun2025any} to enhance both interpretability and diagnostic accuracy.

Building upon these advancements, another significant line of research has focused on leveraging Transformer architectures to model long-range dependencies among patches in WSIs. Despite the effectiveness of self-attention in modeling long-range correlation, the quadratic computational complexity of standard Transformer architectures poses significant challenges when processing tens of thousands of patches in a WSI. To address this limitation, Transformer-based MIL methods could be categorized into two classes, \textit{i.e.}, linear approximation \citep{shao2021transmil} and local-global attention \citep{chen2022scaling,guo2024histgen,li2024longmil}. TransMIL \citep{shao2021transmil}, a pioneering work in linear approximation, adopts the Nystr\"omformer \citep{xiong2021nystromformer} to achieve linear complexity in WSI modeling. However, this approximation strategy often leads to sub-optimal performance due to its inherent limitations in capturing pairwise token interactions, creating an information bottleneck that compromises the model's ability to learn complex dependencies \citep{dao2022flashattention}. In contrast, local-global attention methods like HIPT \citep{chen2022scaling} introduce a hierarchical approach, using a three-level Transformer architecture to progressively aggregate information from cellular features to tissue phenotypes through non-overlapping region slicing. Similarly, HistGen \citep{guo2024histgen}, RRT-MIL \citep{tang2024feature}, and LongMIL \citep{li2024longmil} employ a two-stage strategy, first processing patches within local attention windows before pooling them to enable global attention across the reduced sequence. However, these methods rely on fixed window sizes, which fail to capture the inherently adaptive nature of pathological analysis: malignant regions often require attention to distant but visually similar areas, while normal tissue typically benefits from focusing on local contextual patterns. 

This observation motivates our query-aware dynamic long sequence modeling approach, which adaptively determines relevant regions for each patch based on its pathological characteristics rather than using predetermined fixed attention patterns. Our method efficiently computes region-level metadata through min-max networks to estimate the importance of different tissue regions for each (query) patch. By focusing only on the most relevant regions for each patch, we maintain the expressiveness of full attention while significantly reducing computational cost.

\begin{figure*}[ht]
  \centering
  \includegraphics[width=\textwidth]{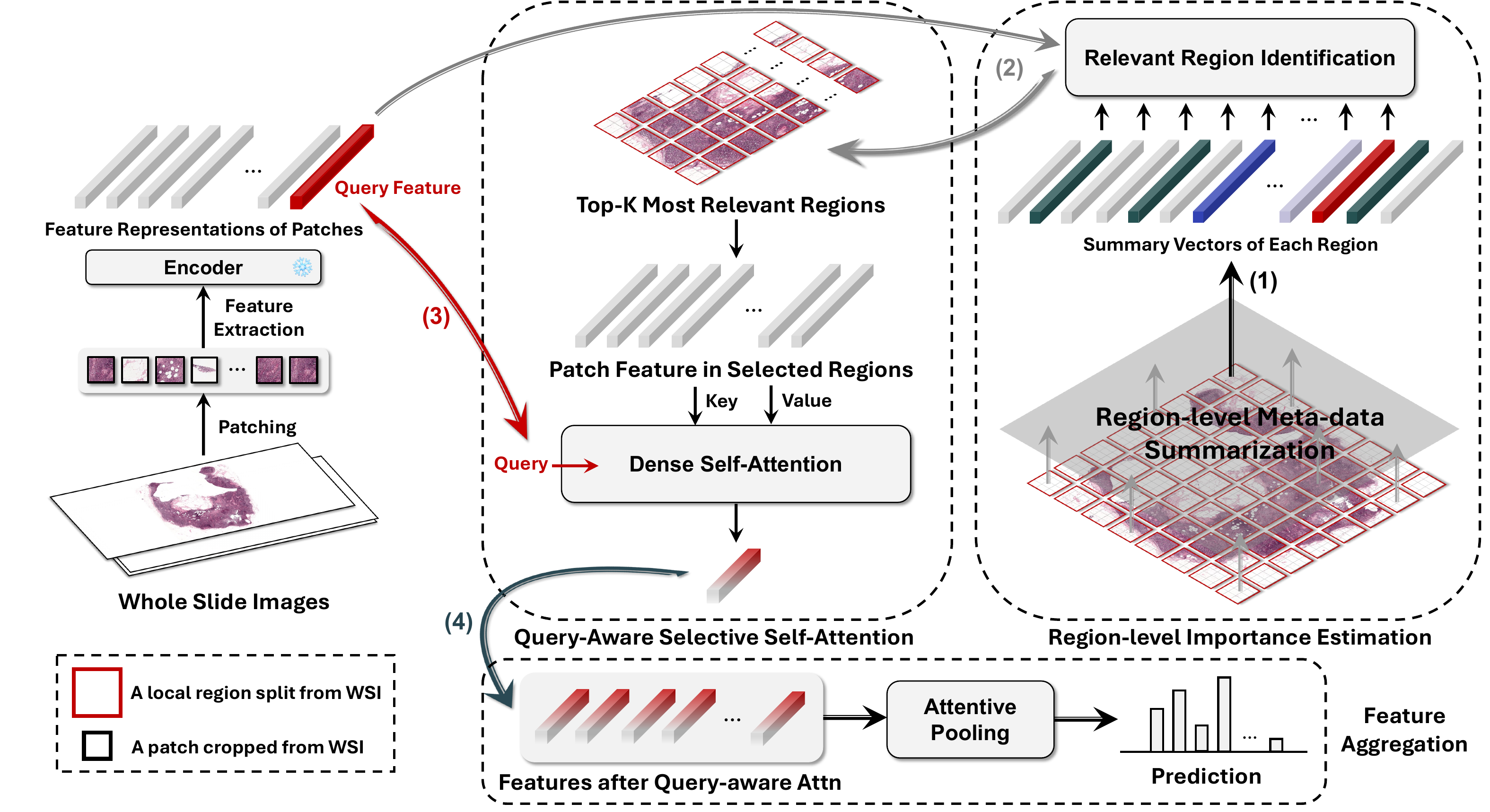}
  \vspace{-0.6cm}
  \caption{Illustration of the proposed \textbf{Querent} framework, which models a WSI via four key steps: (1) region-level metadata summarization from the partitioned WSI, detailed in Fig. \ref{fig:4}, (2) identification of relevant regions for query patches through efficient importance scoring, (3) query-aware selective self-attention computation between query patch and patches in selected regions, and (4) feature aggregation with attentive pooling for final prediction. The framework enables dynamic modeling of long-range contextual relationships in gigapixel WSIs through efficient region relevance identification and query-aware selective attention computation.} 
  \vspace{-0.3cm}
  \label{fig:methodology}
\end{figure*}

\section{Methods}
\subsection{Problem Formulation and Solution Overview}\label{sec:3.1}
\textbf{\textit{Problem Formulation:}} Given a WSI $W$, we first partition it into a set of non-overlapping smaller patches $\{p_1, p_2, ..., p_N\}$ at a fixed magnification level. Each patch is then processed through a pre-trained encoder $\phi(\cdot)$ to obtain feature representations. Specifically, a CPath foundation model named PLIP \citep{huang2023visual} is used for this purpose. Formally, for each patch $p_i$, we obtain its feature representation $x_i = \phi(p_i)$, where $x_i \in \mathbb{R}^{d}$ is a $d$-dimensional feature vector. This results in a bag of instances $X = \{x_1, x_2, ..., x_N\}$ representing the entire WSI. The goal of WSI analysis is to learn a mapping function $f: \mathcal{X} \rightarrow \mathcal{Y}$ that predicts the slide-level label $y$ based on all patches within the WSI through a weakly-supervised manner. The key challenge lies in effectively aggregating information from thousands of patches while capturing both their specific contexts and inter-patch semantic relationships.

\textbf{\textit{Solution Overview:}} Here, we describe the overview of our method comprising 4 major steps as depicted in Fig. \ref{fig:methodology}.

\textbf{Step 1:} For each WSI, we partition the WSI into local regions, where each region contains multiple patches. For each region, we compute region-level metadata through a summarization mechanism. This metadata encapsulates the key characteristics of each region through statistical measures (\textit{e.g.}, min/max features) and serves as a compact representation for importance estimation.

\textbf{Step 2:} Given a query patch, we leverage the pre-computed region-level metadata to identify the top-K most relevant regions through importance scoring efficiently. This enables prioritized processing of regions most likely to contain meaningful contextual information for the current query.

\textbf{Step 3:} Then, we employ query-aware selective attention to process the query with the most relevant regions. This step involves computing dense self-attention between the query and patches within selected regions, allowing for detailed analysis of patch-to-patch relationships while maintaining computational efficiency. The selection is guided by the relevance scores computed using the region-level metadata.

\textbf{Step 4:} Finally, we aggregate the refined features through an attentive pooling mechanism to generate slide-level predictions, which combines the contextually enhanced patch representations while emphasizing the most diagnostically relevant features.

\subsection{Querent for dynamic long sequence WSI modeling}\label{sec:3.2}
\subsubsection{Region-level Metadata Summarization}
Given the extracted patch features $X = \{x_1, x_2, ..., x_N\}$, we first organize them into regions $\mathcal{R} = \{R_1, R_2, ..., R_M\}$, where each region $R_i$ contains a fixed number of $K$ patches. 

\begin{figure}[t]
    \centering
    \includegraphics[width=0.9\linewidth]{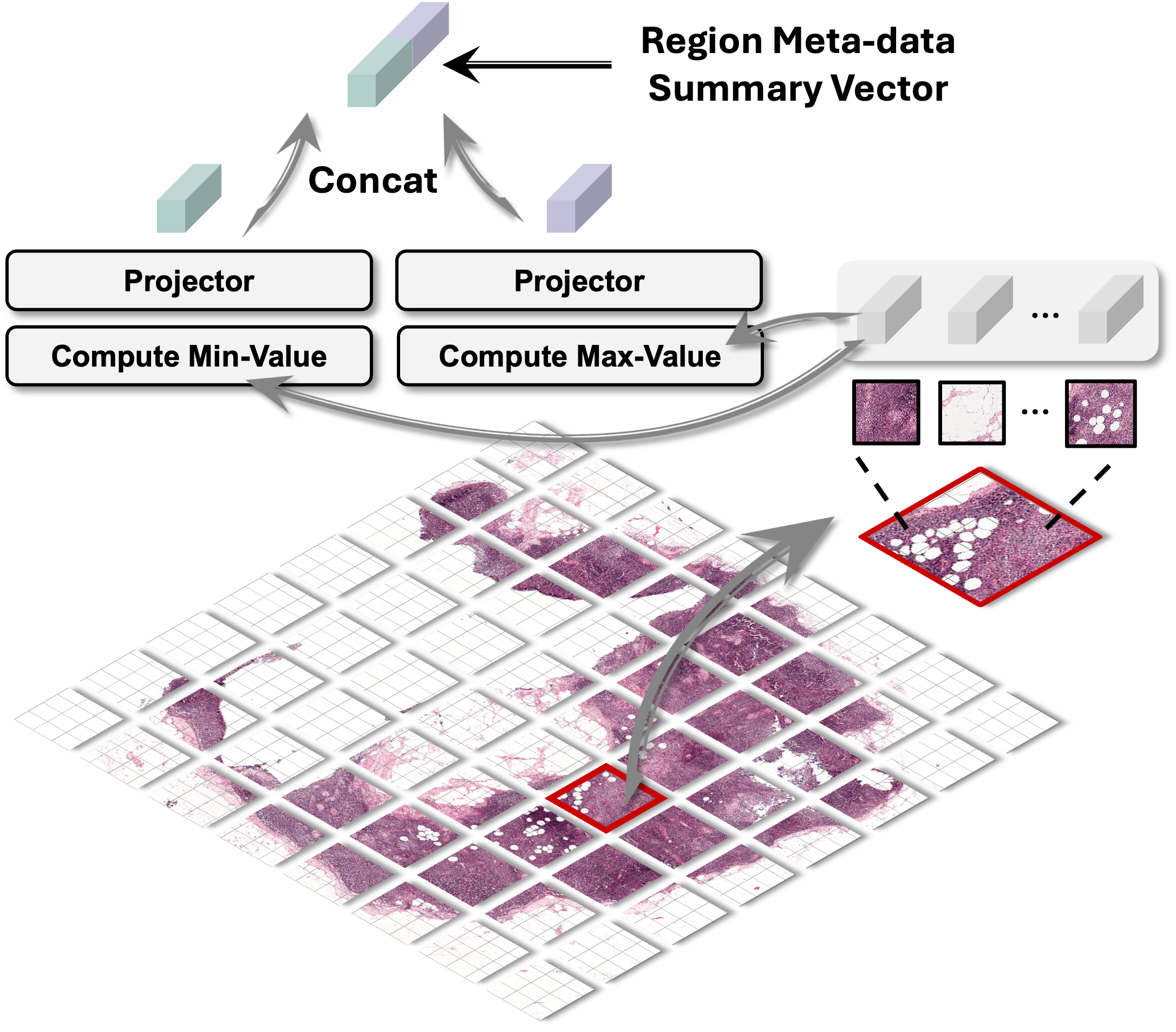}
    \vspace{-0.2cm}
    \caption{Illustration of the region-level metadata summarization process. Each region from the WSI is represented by summary vectors computed from its constituent patches. These summary vectors capture the statistical characteristics (minimum and maximum values) across all patches within each region, providing an efficient representation for subsequent importance estimation.}
    \label{fig:4}
    \vspace{-0.2cm}
\end{figure}

For each region $R_i$, we compute metadata vectors that capture the statistical characteristics of all patches within that region, as illustrated in Fig. \ref{fig:4}. Specifically, we compute two types of summary vectors: minimum and maximum feature values across all patches in the region. Formally, for region $R_i$ containing patches $\{x_{i1}, x_{i2}, ..., x_{iK}\}$, we compute:
\begin{equation}
    m^{\text{min}}_i = \min_{j \in \{1...K\}} x_{ij},~~m^{\text{max}}_i = \max_{j \in \{1...K\}} x_{ij}
\end{equation}
where $m^{\text{min}}_i, m^{\text{max}}_i \in \mathbb{R}^d$ represent the element-wise minimum and maximum values across all patches in region $R_i$. These summary vectors are then transformed through learnable projections:
\begin{equation}
    \hat{m}^{\text{min}}_i = f_{\text{min}}(m^{\text{min}}_i),~~\hat{m}^{\text{max}}_i = f_{\text{max}}(m^{\text{max}}_i)
\end{equation}
where $f_{\text{min}}$ and $f_{\text{max}}$ are neural networks that project the summary vectors into a shared embedding space. This projection allows the metadata to be directly comparable with query features in the subsequent importance estimation step. The resulting region-level metadata provides a compact yet informative representation of each region's content, enabling efficient relevance assessment without requiring exhaustive patch-level computations.

\subsubsection{Region Importance Estimation for Query}
Given a query patch feature $q \in \mathbb{R}^d$, our goal is to efficiently identify the most relevant regions for this query using the pre-computed region-level metadata. For each region $R_i$, we estimate its importance score based on the potential interaction between the query and its metadata vectors.

Specifically, we first project the query feature into the same embedding space as the region metadata $\hat{q} = f_q(q)$, where $f_q$ is a learnable projection network. The importance score for region $R_i$ is then computed by evaluating the maximum possible interaction between the query and the region's metadata bounds:
\begin{equation}
    s_i = \max(|\langle \hat{q}, \hat{m}^{\text{min}}_i \rangle|, |\langle \hat{q}, \hat{m}^{\text{max}}_i \rangle|)
\end{equation}
where $\langle \cdot, \cdot \rangle$ denotes the dot product operation. The importance score $s_i$ provides an upper bound on the potential relevance of any patch within region $R_i$ to the current query. Based on these scores, we select the top-K regions with the highest importance scores:
\begin{equation}
    \mathcal{R}_q = \text{TopK}(\{(R_i, s_i)\}_{i=1}^M)
\end{equation}
This efficient scoring mechanism allows us to identify the most promising regions for detailed attention computation without examining every patch, significantly reducing the computational complexity while maintaining the ability to capture long-range dependencies.

\subsubsection{Query-Aware Selective Attention}
After identifying the relevant regions $\mathcal{R}q$ for a query patch $q$, we compute dense self-attention between the query and patches within these selected regions. For a selected region $R_i \in \mathcal{R}q$, we first compute query ($\mathbf{Q}$), key ($\mathbf{K}$), and value ($\mathbf{V}$) representations through linear projections, followed by attention score computation:
\begin{equation}
[\mathbf{Q}, \mathbf{K}, \mathbf{V}] = \mathbf{W}_{\text{qkv}}(x), \quad \mathbf{A} = \text{softmax}(\frac{\mathbf{QK}^T}{\sqrt{d_h}})
\end{equation}
where $\mathbf{W}_{\text{qkv}} \in \mathbb{R}^{d \times 3d}$ is a learnable parameter matrix, $x$ represents the concatenation of the query patch and patches from the selected region, and $d_h$ is the dimension of each attention head. The output of the attention layer and multi-head attention are computed as:
\begin{equation}
\mathbf{O} = \mathbf{AV}
\end{equation}
\begin{equation}
\mathbf{O}_h = \text{MultiHead}(\mathbf{Q}, \mathbf{K}, \mathbf{V}) = \text{Concat}(\mathbf{O}_1, \ldots, \mathbf{O}_H)\mathbf{W}_O
\end{equation}
where $\mathbf{W}_O \in \mathbb{R}^{Hd_h \times d}$ is a learnable projection matrix. Through this selective attention mechanism, each query patch attends only to patches within the most relevant regions, achieving a balance between computational efficiency and comprehensive contextual modeling. The computational complexity is reduced from $O(N^2)$ to $O(NK)$, where $N$ is the total number of patches and $K$ is the number of patches in selected regions.

The selective attention mechanism described above provides an efficient approximation of full self-attention while maintaining its key properties. To formally characterize the relationship between our query-aware selective attention and standard self-attention, we establish the following theoretical guarantee. This theorem demonstrates that under reasonable conditions regarding the input distribution and model parameters, our selective attention mechanism can approximate full self-attention with bounded error. Specifically, we show:
\begin{theorem}[Query-Aware Attention Approximation]
Let $\mathbf{A}$ be the query-aware attention matrix (Def. \ref{def:query-aware}), and $\mathbf{B}$ be the full self-attention matrix (Def. \ref{def:self-attn}). Assume attention scores decay exponentially with spatial distance: $\exp(-\alpha d(i,j))$ bounds the attention score decay for distance $d(i,j)$. For any input sequence $\mathbf{X}$ with $L$ patches and $d$ dimensions, there exist \textbf{random projection matrices} $\mathbf{W}_Q, \mathbf{W}_K \in \mathbb{R}^{d \times d}$ such that $\|\mathbf{A}-\mathbf{B}\|_{F}\leq\left(2 + \frac{B}{\sqrt{d}}\right)\epsilon$
with probability at least $1 - \delta$, provided:
\begin{enumerate}
    \item The hidden dimension satisfies, and the number of selected regions per query satisfies:
\begin{equation*}
d \geq C_1 \cdot \frac{\log(L/\delta)}{\epsilon^2}, \quad k \geq \frac{C_2}{\alpha} \cdot \log\left(\frac{1}{\epsilon}\right)
\end{equation*}
    where $C_1 = 8B^4$ (from JL inner-product preservation \citep{kaban2015improved}), and $C_2 = 2$.
    \item For each region $R_i$, the diameter satisfies:
\begin{equation*}
\mathrm{diam}(R_i) \leq \min\left(\frac{\epsilon}{L \cdot \sqrt{d}}, \frac{1}{\alpha}\right)
\end{equation*}
    \item Regions are spatially separated such that:
    \begin{equation*}
        \forall i \neq j, \quad d(R_i, R_j) \geq \frac{C_3}{\alpha}
    \end{equation*}
    where $C_3 = \frac{1}{2}$ ensures $\sum_{m=k+1}^\infty e^{-C_3 m} \leq \epsilon$.
\end{enumerate}
\end{theorem}
\begin{proof}
    The detailed derivation of proof can be found in \textbf{Appendix} \ref{appendix:B.1}.
    \vspace{-0.4cm}
\end{proof}

\begingroup
\setlength{\tabcolsep}{4pt} 
\renewcommand{\arraystretch}{1.5} 
\begin{table*}[t]
\centering
\caption{Results of biomarker prediction, gene mutation prediction, and cancer subtyping tasks, with accuracy, AUC, F1 score reported. The best results are in bold, and the second-best results are underlined. Rows in \colorbox{gray!20!white}{gray color} represent Self-Attn-based methods.}
\vspace{0.2cm}
\begin{adjustbox}{width=\textwidth}
\scalebox{0.8}{
\begin{tabular}{cccccccccccc}
\toprule
\toprule
\multicolumn{1}{c}{\multirow{2}{*}{\textbf{Methods}}} & \multicolumn{3}{c}{\textbf{BCNB-ER ($n=1038$)}} &  &\multicolumn{3}{c}{\textbf{TCGA-LUAD TP53 ($n=469$)}} &  & \multicolumn{3}{c}{\textbf{UBC-OCEAN ($n=527$)}} \\ 
\cline{2-4} \cline{6-8} \cline{10-12}
& ACC & AUC & F1 Score & & ACC & AUC & F1 Score & & ACC & AUC & F1 Score\\ 
\midrule
\midrule
Mean Pooling & $0.806_{\pm 0.047}$& $0.820_{\pm 0.044}$& $0.700_{\pm 0.062}$& & $0.639_{\pm 0.061}$ & $0.672_{\pm 0.081}$ & $0.630_{\pm 0.063}$ & & $0.788_{\pm 0.017}$ & $0.941_{\pm 0.017}$ & $0.759_{\pm 0.038}$ \\
Max Pooling & $0.810_{\pm 0.016}$ & $0.819_{\pm 0.051}$& $0.723_{\pm 0.029}$& & $0.659_{\pm 0.049}$ & $0.660_{\pm 0.048}$ & $0.650_{\pm 0.046}$ & & $0.800_{\pm 0.031}$ & \underline{$0.946_{\pm 0.015}$} & $0.782_{\pm 0.033}$ \\
ABMIL  & \underline{$0.825_{\pm 0.038}$} & $0.825_{\pm 0.072}$& $0.726_{\pm 0.085}$& & $0.639_{\pm 0.024}$ & $0.688_{\pm 0.077}$ & $0.630_{\pm 0.017}$  & & \underline{$0.823_{\pm 0.035}$} & $0.942_{\pm 0.019}$ & \underline{$0.796_{\pm 0.039}$} \\
DS-MIL  & $0.817_{\pm 0.050}$& $0.814_{\pm 0.087}$& \underline{$0.734_{\pm 0.077}$}& & $0.629_{\pm 0.047}$ & $0.684_{\pm 0.072}$ & $0.614_{\pm 0.046}$ & & $0.781_{\pm 0.029}$ & $0.938_{\pm 0.018}$ & $0.753_{\pm 0.024}$ \\
CLAM-SB   &$0.821_{\pm 0.052}$ & $0.821_{\pm 0.050}$& $0.732_{\pm 0.070}$& & $0.615_{\pm 0.073}$ & $0.647_{\pm 0.061}$ & $0.607_{\pm 0.071}$  & & $0.808_{\pm 0.021}$ & $0.942_{\pm 0.010}$ & \underline{$0.796_{\pm 0.025}$} \\
DTFD & $0.786_{\pm 0.029}$ & \underline{$0.835_{\pm 0.040}$} & $0.715_{\pm 0.034}$ & & $0.644_{\pm 0.057}$ & $0.680_{\pm 0.045}$ & $0.636_{\pm 0.052}$ & & $0.793_{\pm 0.021}$ & $0.941_{\pm 0.014}$ & $0.779_{\pm 0.029}$ \\
WiKG  & $0.794_{\pm 0.027}$& $0.815_{\pm 0.038}$& $0.717_{\pm 0.026}$& & $0.654_{\pm 0.067}$ & $0.661_{\pm 0.076}$ & $0.642_{\pm 0.048}$ & & $0.819_{\pm 0.065}$ & $0.944_{\pm 0.023}$ & $0.782_{\pm 0.080}$ \\
MambaMIL & \underline{$0.825_{\pm 0.043}$} & $0.820_{\pm 0.067}$ & $0.719_{\pm 0.017}$  & & $0.639_{\pm 0.064}$ & $0.685_{\pm 0.075}$ & $0.627_{\pm 0.079}$ & & $0.808_{\pm 0.021}$ & $0.941_{\pm 0.020}$ & $0.785_{\pm 0.031}$ \\
\hline
\rowcolor{gray!20!white} TransMIL & $0.802_{\pm 0.029}$& $0.780_{\pm 0.086}$& $0.664_{\pm 0.063}$& & $0.615_{\pm 0.028}$ & $0.658_{\pm 0.094}$ & $0.596_{\pm 0.034}$ & & $0.723_{\pm 0.055}$ & $0.928_{\pm 0.017}$ & $0.698_{\pm 0.051}$ \\
\rowcolor{gray!20!white} HIPT & $0.796_{\pm 0.016}$ & $0.758_{\pm 0.056}$ & $0.681_{\pm 0.034}$ & & $0.600_{\pm 0.029}$ & $0.672_{\pm 0.062}$ & $0.588_{\pm 0.031}$ & & $0.781_{\pm 0.040}$ & $0.913_{\pm 0.035}$ & $0.737_{\pm 0.068}$ \\
\rowcolor{gray!20!white} HistGen & $0.791_{\pm 0.027}$ & $0.790_{\pm 0.066}$ & $0.695_{\pm 0.049}$  & & \underline{$0.668_{\pm 0.061}$} & $0.665_{\pm 0.056}$ & \underline{$0.658_{\pm 0.063}$} & &$0.784_{\pm 0.025}$  & $0.934_{\pm 0.009}$ & $0.764_{\pm 0.053}$ \\
\rowcolor{gray!20!white} RRT-MIL & $0.812_{\pm 0.052}$ & $0.814_{\pm 0.054}$ & $0.688_{\pm 0.085}$ & & $0.600_{\pm 0.077}$ & $0.657_{\pm 0.029}$ & $0.581_{\pm 0.077}$ & & $0.804_{\pm 0.022}$ & $0.939_{\pm 0.016}$ & $0.783_{\pm 0.039}$ \\
\rowcolor{gray!20!white} LongMIL & $0.781_{\pm 0.018}$ & $0.782_{\pm 0.058}$ & $0.653_{\pm 0.029}$ & & $0.663_{\pm 0.081}$ & \underline{$0.693_{\pm 0.086}$} & $0.657_{\pm 0.079}$ & & $0.760_{\pm 0.022}$ & $0.921_{\pm 0.016}$ & $0.742_{\pm 0.033}$ \\
\rowcolor{gray!20!white} \textbf{Querent (Ours)} &  \pmb{$0.836_{\pm 0.043}$} & \pmb{$0.848_{\pm 0.042}$} & \pmb{$0.739_{\pm 0.042}$} & & \pmb{$0.678_{\pm 0.068}$} & \pmb{$0.706_{\pm 0.090}$} & \pmb{$0.672_{\pm 0.070}$} & & \pmb{$0.835_{\pm 0.015}$} & \pmb{$0.956_{\pm 0.019}$} & \pmb{$0.806_{\pm 0.041}$} \\
\bottomrule
\bottomrule
\end{tabular}}
\end{adjustbox}
\label{tab:classification}
\vspace{-0.3cm}
\end{table*}
\endgroup
\begingroup
\setlength{\tabcolsep}{4pt} 
\renewcommand{\arraystretch}{1.5} 
\begin{table*}[t]
\centering
\caption{Results of survival prediction on 8 TCGA subsets with C-Index score reported. The best results are in bold, and the second-best results are underlined. Rows in \colorbox{gray!20!white}{gray color} represent Self-Attn-based methods.}
\vspace{0.2cm}
\begin{adjustbox}{width=\textwidth}
\scalebox{0.8}{
\begin{tabular}{cccccccccc}
\toprule
\toprule
\textbf{Methods} & BRCA  &  UCEC &  STAD & LUAD & LUSC  & SKCM  & KIRC & KIRP & Avg. ($\uparrow$) \\
\midrule
\midrule
Mean Pooling & $0.669_{\pm 0.013}$ & \underline{$0.685_{\pm 0.053}$} & $0.594_{\pm 0.044}$  & $0.558_{\pm 0.074}$ & $0.700_{\pm 0.054}$  &  $0.661_{\pm 0.045}$ & $0.616_{\pm 0.066}$ & $0.643_{\pm 0.041}$ & $0.641$\\
Max Pooling & $0.674_{\pm 0.037}$ & $0.649_{\pm 0.037}$  &  $0.539_{\pm 0.054}$ &  $0.519_{\pm 0.065}$& $0.672_{\pm 0.039}$ &   $0.666_{\pm 0.064}$ & $0.581_{\pm 0.031}$ &  $0.593_{\pm 0.024}$ & $0.612$\\
ABMIL  & $0.698_{\pm 0.020}$  & $0.666_{\pm 0.039}$ &  $0.598_{\pm 0.075}$  & $0.581_{\pm 0.063}$ & $0.707_{\pm 0.042}$ & $0.663_{\pm 0.084}$   & $0.606_{\pm 0.065}$ & $0.608_{\pm 0.071}$ & $0.641$ \\
DS-MIL & $0.678_{\pm 0.023}$ & $0.608_{\pm 0.050}$  & $0.547_{\pm 0.073}$  & $0.568_{\pm 0.062}$ & $0.705_{\pm 0.043}$  & \underline{$0.671_{\pm 0.045}$} & $0.582_{\pm 0.079}$ & $0.593_{\pm 0.081}$ &  $0.619$ \\
DTFD & $0.695_{\pm 0.022}$ & $0.672_{\pm 0.032}$ & $0.597_{\pm 0.090}$ & $0.570_{\pm 0.054}$ & \underline{$0.716_{\pm 0.046}$} & $0.649_{\pm 0.048}$   & \pmb{$0.631_{\pm 0.042}$} & $0.598_{\pm 0.023}$ & $0.641$ \\
WiKG  & $0.669_{\pm 0.019}$ & $0.678_{\pm 0.049}$ & $0.612_{\pm 0.066}$  & $0.588_{\pm 0.056}$ & $0.711_{\pm 0.030}$ &   $0.654_{\pm 0.083}$ & $0.610_{\pm 0.031}$ & \underline{$0.651_{\pm 0.036}$} & \underline{$0.647$} \\
MambaMIL & $0.694_{\pm 0.012}$ & $0.668_{\pm 0.022}$  & $0.590_{\pm 0.039}$ & \pmb{$0.609_{\pm 0.059}$} & $0.692_{\pm 0.044}$ &   $0.666_{\pm 0.046}$ & $0.614_{\pm 0.067}$ &  $0.641_{\pm 0.042}$ & \underline{$0.647$}\\
\hline
\rowcolor{gray!20!white} TransMIL & $0.643_{\pm 0.038}$ & $0.642_{\pm 0.049}$  & $0.568_{\pm 0.082}$  & $0.498_{\pm 0.067}$ & $0.657_{\pm 0.056}$  & $0.586_{\pm 0.042}$ & $0.602_{\pm 0.043}$ & $0.567_{\pm 0.092}$  & $0.595$ \\
\rowcolor{gray!20!white} HIPT & \underline{$0.717_{\pm 0.006}$}  &  $0.647_{\pm 0.031}$  &  $0.588_{\pm 0.065}$  & $0.536_{\pm 0.037}$ &  $0.603_{\pm 0.054}$ & $0.616_{\pm 0.039}$ & $0.578_{\pm 0.036}$  & $0.559_{\pm 0.103}$  & $0.606$ \\
\rowcolor{gray!20!white} HistGen & $0.688_{\pm 0.041}$ & $0.664_{\pm 0.041}$ & \underline{$0.619_{\pm 0.064}$}  & $0.562_{\pm 0.068}$ & $0.708_{\pm 0.026}$   & $0.611_{\pm 0.057}$ & $0.610_{\pm 0.049}$ & $0.560_{\pm 0.055}$  & $0.628$ \\
\rowcolor{gray!20!white} RRT-MIL & $0.700_{\pm 0.022}$ & $0.672_{\pm 0.029}$ & $0.555_{\pm 0.064}$ & $0.580_{\pm 0.089}$  & $0.669_{\pm 0.070}$ & $0.652_{\pm 0.063}$  &  $0.592_{\pm 0.048}$ &  $0.561_{\pm 0.041}$ & $0.623$ \\
\rowcolor{gray!20!white} LongMIL & $0.652_{\pm 0.024}$ & $0.633_{\pm 0.062}$ & $0.591_{\pm 0.094}$  & $0.535_{\pm 0.038}$ & $0.674_{\pm0.074}$   & $0.658_{\pm 0.026}$ & \pmb{$0.631_{\pm 0.031}$} & $0.626_{\pm 0.055}$ & $0.625$\\
\rowcolor{gray!20!white} \textbf{Querent (Ours)} & \pmb{$0.720_{\pm 0.012}$} &  \pmb{$0.691_{\pm 0.062}$} & \pmb{$0.636_{\pm 0.028}$} & \underline{$0.608_{\pm 0.050}$} & \pmb{$0.732_{\pm 0.053}$}   & \pmb{$0.682_{\pm 0.045}$} & \underline{$0.626_{\pm 0.038}$} & \pmb{$0.667_{\pm 0.024}$} & \pmb{$0.670$} \\
\bottomrule
\bottomrule
\end{tabular}}
\end{adjustbox}
\label{tab:survival}
\vspace{-0.3cm}
\end{table*}
\endgroup

\subsubsection{Attentive Feature Aggregation}
After obtaining contextually enhanced representations for all patches through query-aware selective attention, we employ an attentive pooling mechanism to aggregate these features for slide-level prediction. Given the refined patch features $\{x'_1, x'_2, ..., x'_N\}$, we compute attention weights through a learnable attention network:
\begin{equation}
w_i = \sigma(f_a(x'_i)), \quad a_i = \frac{\exp(w_i)}{\sum_{j=1}^N \exp(w_j)}
\end{equation}
where $f_a$ is a multi-layer perceptron that produces a scalar score for each patch, and $\sigma$ is the sigmoid activation function. The attention weights $a_i$ are normalized through softmax to ensure they sum to one. The final slide-level representation is computed as the weighted sum of all patch features, which is then passed through a final classification layer:
\begin{equation}
z = \sum_{i=1}^N a_i x'_i, \quad \hat{y} = f_c(z)
\end{equation}
where $f_c$ is a fully connected layer that outputs the predicted class probabilities. This attentive pooling mechanism allows the model to emphasize diagnostically relevant patches while suppressing the influence of irrelevant or background regions in the final prediction.

Our model is trained end-to-end using task-appropriate loss functions. For classification tasks, we employ cross-entropy (CE) loss $\mathcal{L}_{\text{ce}}$. For survival prediction tasks, we utilize the negative log-likelihood (NLL) survival loss $\mathcal{L}_{\text{surv}}$ \citep{cox1972regression,katzman2018deepsurv,zadeh2020bias}. The detailed formulations of these loss functions and their optimization procedures are provided in \textbf{Appendix} \ref{appendix:loss}.

\section{Experiments}
\subsection{Tasks and Datasets}
We evaluate Querent across four types of tasks over 11 publicly available datasets (see \textbf{Appendix} \ref{appendix:dataset} for details):

\textbf{(1) Biomarker prediction.} BCNB \citep{xu2021predicting} ($n=1038$), a dataset of early breast cancer core-needle biopsy WSI is used to predict tumor clinical characteristic estrogen receptor (ER). 

\textbf{(2) Gene mutation prediction.} A subset of The Cancer Genome Atlas (TCGA) \citep{tomczak2015review} is used (TCGA-LUAD, $n=469$) to predict TP53 gene mutation from lung adenocarcinoma WSIs. 

\textbf{(3) Cancer subtyping.} UBC-OCEAN ($n=527$), an ovarian cancer dataset, is used to predict 5 cancer subtypes.

\textbf{(4) Survival Analysis.} 8 TCGA subsets are used for this task, including TCGA-BRCA ($n=1025$), TCGA-UCEC ($n=497$), TCGA-STAD ($n=365$), TCGA-LUAD ($n=457$), TCGA-LUSC ($n=454$), TCGA-SKCM ($n=417$), TCGA-KIRC ($n=500$), and TCGA-KIRP ($n=263$). 

These four tasks represent key challenges across the computational pathology pipeline: from molecular-level analysis (biomarker/mutation prediction) to clinical assessment (subtyping) and patient outcomes (survival). 

We use 5-fold cross-validation for model training and evaluation and report the results' mean and standard deviation. For classification tasks, accuracy (ACC), area under the curve (AUC), and F1 Score are reported for comparison. For survival analysis, the concordance index (C-Index) \citep{harrell1982evaluating} is used for evaluation.

\subsection{Baselines}
We employ SOTA models including Max/Mean Pooling, ABMIL \citep{ilse2018attention}, DS-MIL \citep{li2021dual}, DTFD \citep{zhang2022dtfd}, WiKG \citep{li2024dynamic}, and MambaMIL \citep{yang2024mambamil} for comparison. For Transformer-based MIL models, we involve TransMIL \citep{shao2021transmil}, HIPT \citep{chen2022scaling}, HistGen \citep{guo2024histgen}, RRT-MIL \citep{tang2024feature}, and LongMIL \citep{li2024longmil}. For all models in comparison, we use PLIP \citep{huang2023visual} as the patch feature encoder. Further information on all baselines and implementation details are available in \textbf{Appendix} \ref{appendix:baseline} and \textbf{Appendix} \ref{appendix:implementation}, respectively.

\vspace{-0.2cm}
\section{Results}
\vspace{-0.1cm}
\subsection{Comparison to SOTA Methods}
Tab. \ref{tab:classification} shows the evaluation results on classification tasks, including biomarker prediction (BCNB-ER), gene mutation prediction (TCGA-LUAD TP53), and cancer subtyping (UBC-OCEAN). Across all three tasks, our method consistently achieves state-of-the-art performance. For BCNB-ER, Querent achieves an accuracy of $0.836$, AUC of $0.848$, and F1 score of $0.739$, outperforming all MIL baselines and transformer-based counterparts. On TCGA-LUAD TP53, our method demonstrates superior performance with an accuracy of $0.678$ and AUC of $0.706$, showing particular advantages over other transformer-based methods. For the complex task of 5-class cancer subtyping on UBC-OCEAN, Querent achieves significant improvements with an accuracy of $0.835$ and AUC of $0.956$, surpassing the second-best method by 1.2\% and 1.0\%, respectively.

Moreover, Tab. \ref{tab:survival} demonstrates the survival prediction results across eight TCGA cancer types. Querent achieves the highest average C-Index of $0.670$, representing a substantial improvement over the second-best methods (WiKG and MambaMIL, $0.647$). The superior performance is consistent across different cancer types, with our method achieving the best results on BRCA ($0.720$), UCEC ($0.691$), STAD ($0.636$), LUSC ($0.732$), SKCM ($0.682$), and KIRP ($0.667$). This consistent improvement across diverse cancer types demonstrates the robustness and generalizability of our method in capturing prognostic features from WSIs. 

\subsection{Ablation Study}
\begin{figure}[t]
    \centering
    \includegraphics[width=0.98\linewidth]{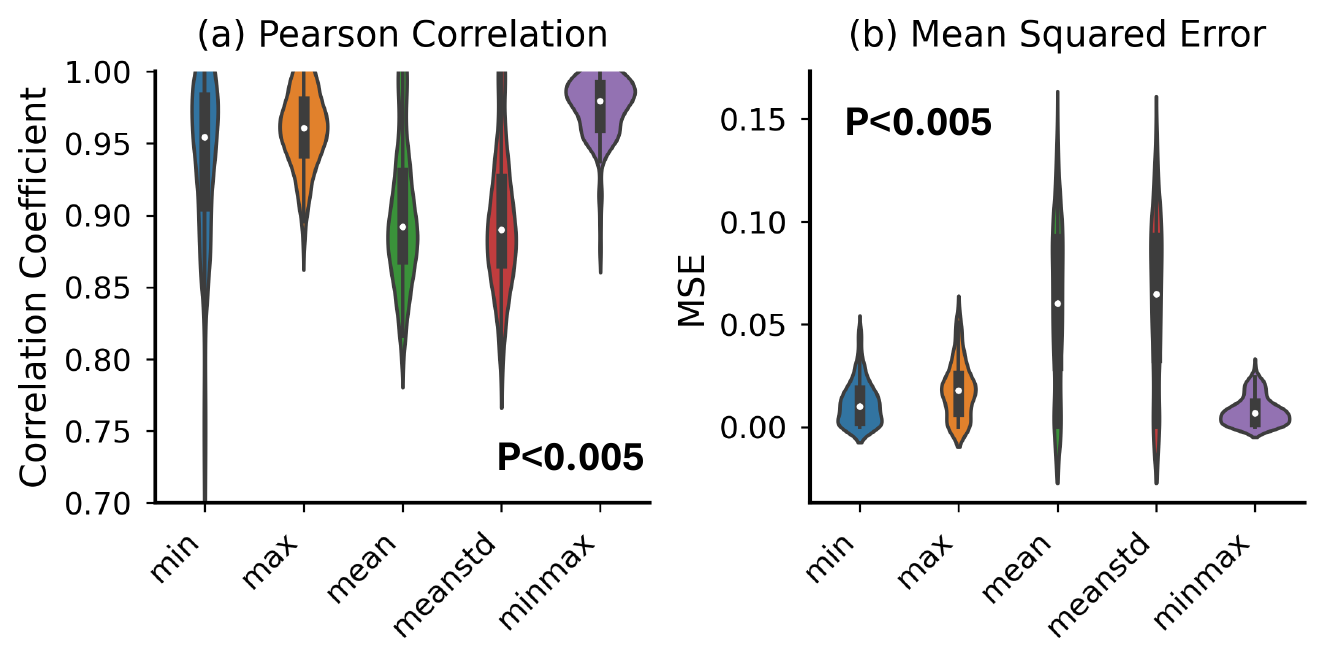}
    \vspace{-0.4cm}
    \caption{Ablation on \textbf{Querent} using min, max, mean, and mean $\pm$std strategies compared to our min-max method on TCGA-LUAD TP53 gene mutation dataset (details in \textbf{Appendix} \ref{appendix:ablation1}).}
    \label{fig:ablation_summarization}
    \vspace{-0.3cm}
\end{figure}

\textbf{Region-level Metadata Summarization Strategy.} We evaluate the effectiveness of different region-level feature summarization strategies by comparing pairwise distance relationships between regions before and after summarization. As shown in Fig. \ref{fig:ablation_summarization}, our proposed min-max summarization strategy demonstrates superior performance as measured by both the Pearson correlation coefficient and mean squared error (MSE).
Specifically, our min-max approach achieves the highest correlation (0.975) and lowest MSE (0.008), significantly outperforming ($p<0.005$) alternative methods including individual min (correlation: 0.937, MSE: 0.012) or max summarization (correlation: 0.959, MSE: 0.018), mean (correlation: 0.902, MSE: 0.058), and mean-std approaches (correlation: 0.897, MSE: 0.062), indicating that combining both minimum and maximum values effectively preserves the structural relationships between regions.

\begingroup
\setlength{\tabcolsep}{4pt} 
\renewcommand{\arraystretch}{1.5} 
\vspace{-0.2cm}
\begin{table}[ht]
\centering
\caption{Ablation on the importance estimation module of \textbf{Querent}, with reported results on TCGA-LUAD TP53 mutation and UBC-OCEAN ovarian cancer datasets (details in \textbf{Appendix} \ref{appendix:ablation2}).}
\vspace{0.2cm}
\begin{adjustbox}{width=8.25cm}
\scalebox{0.8}{
\begin{tabular}{l|ccc}
\toprule
 \textbf{Importance Estimation} & Accuracy & AUC & F1 Score \\
\midrule
\multicolumn{4}{c}{\textbf{TCGA-LUAD TP53 Gene Mutation Prediction}} \\
\midrule
Estimation Side Network & $0.580_{\pm 0.076}$ & $0.660_{\pm 0.042}$ & $0.568_{\pm 0.073}$  \\
Random Region Selection& $0.649_{\pm 0.063}$  & $0.686_{\pm 0.071}$  & $0.643_{\pm 0.061}$  \\
\textbf{Querent (Ours)} & \pmb{$0.678_{\pm 0.068}$} & \pmb{$0.706_{\pm 0.090}$} & \pmb{$0.672_{\pm 0.070}$} \\
\midrule
\multicolumn{4}{c}{\textbf{UBC-OCEAN Ovarian Cancer Subtyping}} \\
\midrule
Estimation Side Network &  $0.731_{\pm 0.036}$ & $0.903_{\pm 0.019}$  &  $0.690_{\pm 0.026}$ \\
Random Region Selection & $0.746_{\pm 0.056}$  & $0.914_{\pm 0.036}$  & $0.697_{\pm 0.062}$  \\
\textbf{Querent (Ours)} & \pmb{$0.835_{\pm 0.015}$} & \pmb{$0.956_{\pm 0.019}$} & \pmb{$0.806_{\pm 0.041}$} \\
\bottomrule
\end{tabular}}
\end{adjustbox}
\label{tab:ablation_importance_estimation}
\end{table}
\endgroup
\textbf{Region Importance Estimation Strategy.} Meanwhile, we conduct an ablation study comparing our proposed region importance estimation module against random region selection and an estimation side network, reported in Tab. \ref{tab:ablation_importance_estimation}. Our approach consistently outperforms both baselines across TCGA-LUAD and UBC-OCEAN datasets. For TP53 mutation prediction, our method achieves 0.678 accuracy and 0.706 AUC, showing moderate improvements over the baselines. The gains are more substantial in ovarian cancer subtyping (UBC-OCEAN), where our approach reaches 0.835 accuracy and 0.956 AUC, representing absolute improvements of 8.9\% in accuracy and 4.2\% in AUC over random selection, demonstrating the effectiveness of our region importance estimation strategy.

\begin{figure}[t]
    \centering
    \includegraphics[width=1\linewidth]{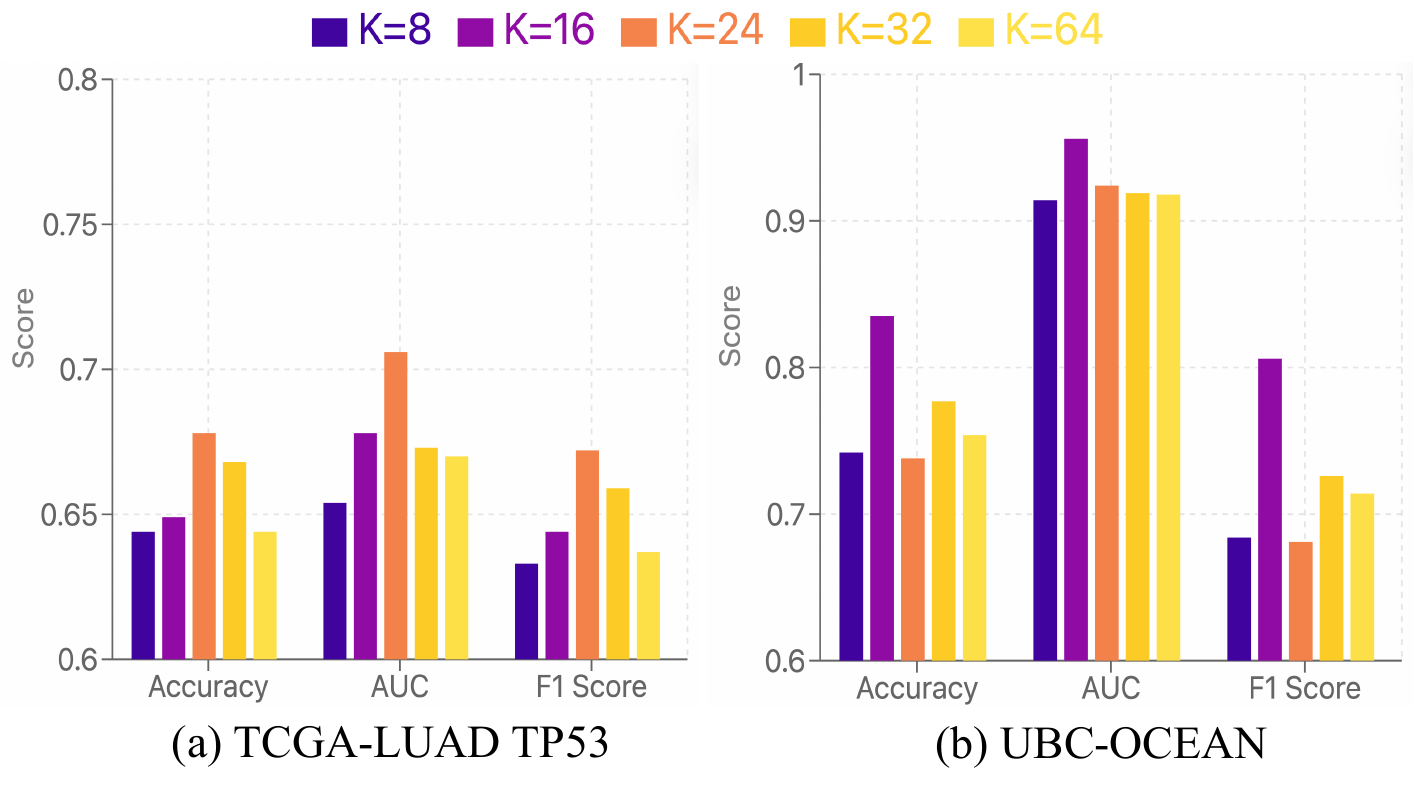}
    \vspace{-0.6cm}
    \caption{Ablation on \textbf{Querent} using different region size $K$, with reported results on TCGA-LUAD for TP53 gene mutation prediction and UBC-OCEAN for ovarian cancer subtyping.}
    \label{fig:ablation_region_size}
    \vspace{-0.1cm}
\end{figure}

\textbf{Region size in Querent.} Fig. \ref{fig:ablation_region_size} shows the impact of applying different region size for our Querent model during the region metadata summarization process. For TP53 mutation prediction in TCGA-LUAD, moderate region sizes (K=24) yielded optimal results with an AUC of 0.706, while both smaller (K=8) and larger (K=64) regions showed decreased performance. In UBC-OCEAN ovarian cancer subtyping, K=16 emerged as the clear optimal choice, achieving the highest accuracy (0.835) and AUC (0.956), with performance gradually declining as region size increased. These results demonstrate that moderate-sized regions are most effective for both tasks. This aligns with pathological intuition, as larger regions may dilute the discriminative local tissue patterns by aggregating potentially heterogeneous areas, while smaller regions might miss important contextual information.
\begin{figure}[t]
    \centering
    \includegraphics[width=1\linewidth]{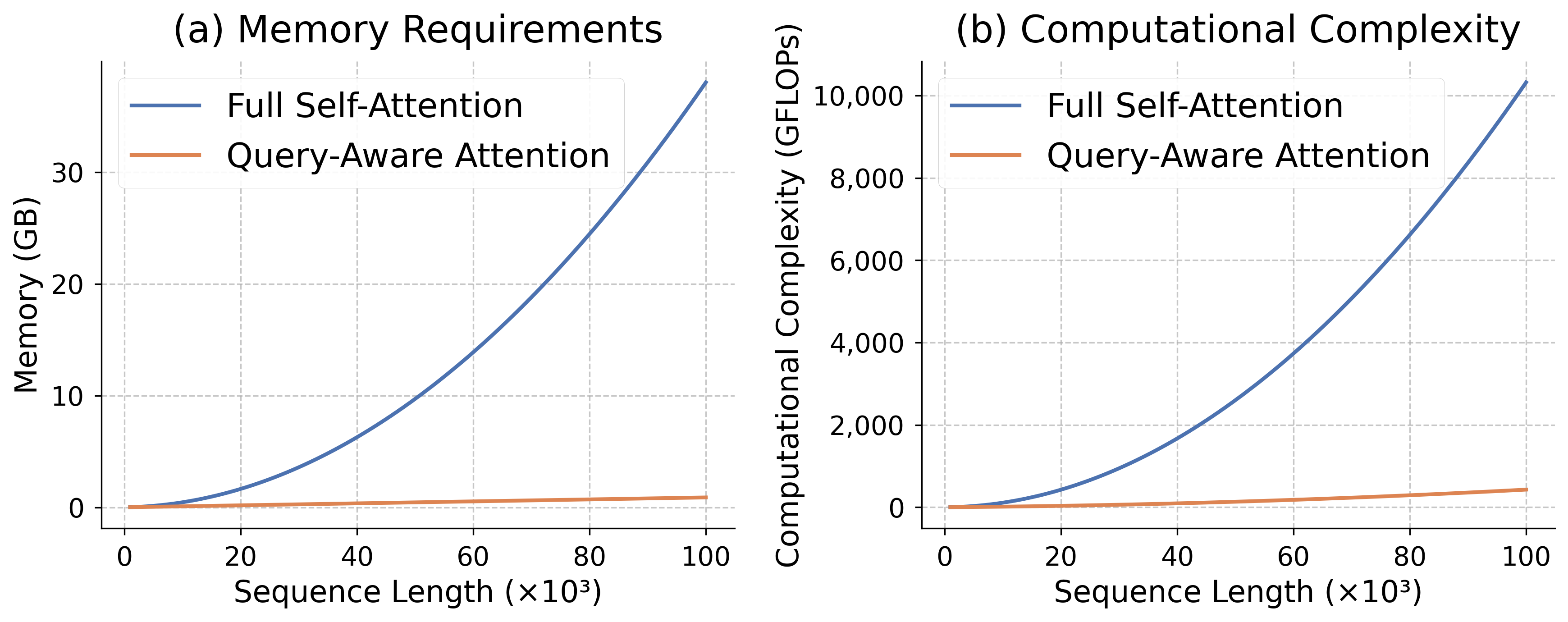}
    \vspace{-0.6cm}
    \caption{Computational efficiency comparison between full self-attention and our query-aware approach. (a) Memory requirements in gigabytes and (b) computational complexity in GFLOPs across different sequence lengths. See detailed analysis in \textbf{Appendix} \ref{appendix:complexity}.}
    \label{fig:complexity}
    \vspace{-0.3cm}
\end{figure}
\subsection{Computational Efficiency} We analyze the computational efficiency of our query-aware attention mechanism compared to full self-attention (illustrated in Fig. \ref{fig:complexity}). 
While full self-attention reaches 37GB memory and 10,000 GFLOPs at 100k sequence length with quadratic scaling, our approach achieves near-linear scaling, requiring only 1GB memory and 500 GFLOPs for a 100k patch sequence. This significant reduction in computational overhead enables our model to efficiently process longer sequences, which is crucial for gigapixel whole-slide image analysis. 

\section{Conclusion}
In this study, we present \textbf{Querent}, a query-aware dynamic modeling framework for understanding long-range contextual correlations in gigapixel WSIs. Inspired by pathological observations, our method aims to dynamically adapt to the unique context of each patch while remaining computationally efficient through region-level metadata summarization and importance-based attention.
Extensive experiments across multiple CPath tasks demonstrate the superior performance of our model compared to state-of-the-art approaches, establishing its effectiveness in whole slide image analysis.






\section*{Acknowledgements}

This work was supported by the National Natural Science Foundation of China (No. 62202403), Hong Kong Innovation and Technology Commission (Project No. MHP/002/22 and ITCPD/17-9), and Research Grants Council of the Hong Kong Special Administrative Region, China (Project No. R6003-22 and C4024-22GF).


\section*{Impact Statement}
This work advances computational pathology through a novel query-aware dynamic long contextual modeling framework for analyzing gigapixel whole slide images. Our research utilizes publicly available pathology datasets with appropriate institutional approvals. While our work demonstrates potential for improving the efficacy and efficiency of histopathological analysis, it is currently intended for research purposes only. Clinical implementation would require extensive external validation and regulatory approval. Our methodology and findings are shared to advance scientific understanding in computational pathology rather than for immediate clinical application.

\bibliography{example_paper}
\bibliographystyle{icml2025}

\newpage
\appendix
\onecolumn
\section{Algorithm Pseudo Code of Querent}
\vspace{-0.2cm}
\begin{algorithm*}[h]
  \caption{Querent: Query-Aware Dynamic Long Sequence Modeling for Gigapixel WSI Analysis}
  \label{alg:querent}
\begin{algorithmic}[1]
  \STATE {\bfseries Input:} Patch features $X = \{x_1, x_2, ..., x_N\}$, region size $K$, number of regions $M$
  \STATE {\bfseries Output:} Slide-level prediction $\hat{y}$ and attention weights $\alpha$
  \STATE {\bfseries Parameter:} Query projection $f_q$, metadata projections $f_{\text{min}}, f_{\text{max}}$, attention $\mathbf{W}_{\text{qkv}}, \mathbf{W}_O$, classifier $f_c$
  
  \STATE // Phase 1: Region-level Metadata Summarization
  \FOR{$i = 1$ to $M$}
    \STATE Extract region patches: $R_i = \{x_{i1}, x_{i2}, ..., x_{iK}\}$
    \STATE Compute min/max metadata: $m^{\text{min}}_i = \min_{j \in \{1...K\}} x_{ij}$, $m^{\text{max}}_i = \max_{j \in \{1...K\}} x_{ij}$
    \STATE Project metadata: $\hat{m}^{\text{min}}_i = f_{\text{min}}(m^{\text{min}}_i)$, $\hat{m}^{\text{max}}_i = f_{\text{max}}(m^{\text{max}}_i)$
  \ENDFOR
  
  \STATE // Phase 2: Region Importance Estimation
  \FOR{each query patch $q$}
    \STATE Project query: $\hat{q} = f_q(q)$
    \FOR{$i = 1$ to $M$}
      \STATE Compute importance: $s_i = \max(|\langle \hat{q}, \hat{m}^{\text{min}}_i \rangle|, |\langle \hat{q}, \hat{m}^{\text{max}}_i \rangle|)$
    \ENDFOR
    \STATE Select top regions: $\mathcal{R}_q = \text{TopK}(\{(R_i, s_i)\}_{i=1}^M)$
  \ENDFOR
  
  \STATE // Phase 3: Query-Aware Selective Attention
  \FOR{each selected region $R_i \in \mathcal{R}_q$}
    \STATE Compute QKV: $[\mathbf{Q}, \mathbf{K}, \mathbf{V}] = \mathbf{W}_{\text{qkv}}(x)$
    \STATE Compute attention scores: $\mathbf{A} = \text{softmax}(\frac{\mathbf{QK}^T}{\sqrt{d_h}})$
    \STATE Compute attention output: $\mathbf{O} = \mathbf{AV}$
    \STATE Multi-head attention: $\mathbf{O}_h = \text{Concat}(\mathbf{O}_1, \ldots, \mathbf{O}_H)\mathbf{W}_O$
  \ENDFOR
  
  \STATE // Phase 4: Attentive Feature Aggregation
  \STATE Compute attention weights: $w_i = \sigma(f_a(x'_i))$
  \STATE Normalize weights:  $a_i = \frac{\exp(w_i)}{\sum_{j=1}^N \exp(w_j)}$
  \STATE Aggregate features: $z = \sum_{i=1}^N a_i x'_i$
  \STATE Compute final prediction: $\hat{y} = f_c(z)$
  
  \STATE \textbf{Return} $\hat{y}$, $\alpha$
\end{algorithmic}
\end{algorithm*}
\vspace{-0.4cm}

\section{Theoretical Analysis}
\subsection{Query-Aware Sparse Attention: A Theoretically Bounded Approximation of Full Self-Attention}\label{appendix:B.1}
\subsubsection{Definitions and Preliminaries}
\begin{definition}[Full Self-Attention Matrix]\label{def:self-attn}
Let \(\mathbf{X}\in\mathbb{R}^{L\times d}\) be the feature matrix of a sequence of patches, where \(L\) is the sequence length and \(d\) is the feature dimension. The full self-attention matrix \(\mathbf{B}\) is defined as:
\[
\mathbf{B}_{i,j}=\exp\left(\frac{\mathbf{q}_{i}^{\top}\mathbf{K}_{j}}{\sqrt{d}}\right)
\]
where \(\mathbf{q}_{i}\) is the query vector for patch \(i\) and \(\mathbf{K}_{j}\) is the key vector for patch \(j\).
\end{definition}

\begin{definition}[Query-Aware Attention Matrix]\label{def:query-aware}
Let \(\mathbf{X}\in\mathbb{R}^{L\times d}\) be the feature matrix of a sequence of patches, where \(L\) is the sequence length and \(d\) is the feature dimension. The query-aware attention matrix \(\mathbf{A}\) is defined as:
\[
\mathbf{A}_{i,j}=\begin{cases}
\exp\left(\frac{\mathbf{q}_{i}^{\top}\mathbf{K}_{j}}{\sqrt{d}}\right) & \text{if } j\in\mathcal{R}_{i} \\
0 & \text{otherwise}
\end{cases}
\]
where \(\mathcal{R}_{i}\) is the set of top-K relevant regions for query patch \(\mathbf{q}_{i}\), and \(\mathbf{K}_{j}\) is the key vector for patch \(j\).
\end{definition}

\begin{definition}[Region-Level Metadata]
For each region \(R_{i}\), the metadata consists of summary statistics computed from the patches within the region. The primary statistics are the minimum and maximum feature vectors:
\[
\mathbf{m}_{\min}^{i}=\min_{j\in R_{i}}\mathbf{x}_{j}, \quad \mathbf{m}_{\max}^{i}=\max_{j\in R_{i}}\mathbf{x}_{j}
\]
These vectors are projected via Lipschitz-continuous neural networks \(f_{\min}\) and \(f_{\max}\) into a shared embedding space:
\[
\hat{\mathbf{m}}_{\min}^{i}=f_{\min}(\mathbf{m}_{\min}^{i}), \quad \hat{\mathbf{m}}_{\max}^{i}=f_{\max}(\mathbf{m}_{\max}^{i})
\]
where \(f_{\min}\) and \(f_{\max}\) are \(L\)-Lipschitz continuous with constant \(L\).
\end{definition}

\subsubsection{Technical Lemmas}

\begin{lemma}[Region Metadata Approximation]\label{lemma1}
Let \(B\) be a bound on the input norms \(\|\mathbf{q}_{i}\|, \|\mathbf{K}_{j}\|\leq B\). For any query patch \(\mathbf{q}\) and region \(R_{i}\), assuming \(L\)-Lipschitz continuous projection functions, the interaction scores satisfy:
\[
\max_{j\in R_{i}}|\langle\mathbf{q}, \mathbf{x}_{j}\rangle - \langle\hat{\mathbf{q}}, \hat{\mathbf{m}}_{\min}^{i}\rangle| \leq B \cdot L \cdot \mathrm{diam}(R_{i}) = \epsilon_{1}
\]
and similarly for \(\hat{\mathbf{m}}_{\max}^{i}\), where \(\mathrm{diam}(R_{i})\) is the diameter of region \(R_{i}\) in feature space.

\begin{proof}
Fix any patch \(\mathbf{x}_{j}\in R_{i}\). By construction:
\[
\mathbf{m}_{\min}^{i} \preceq \mathbf{x}_{j} \preceq \mathbf{m}_{\max}^{i}
\]
where \(\preceq\) denotes element-wise comparison. By Lipschitz continuity:
\[
\|\hat{\mathbf{m}}_{\min}^{i} - f_{\min}(\mathbf{x}_{j})\| \leq L \|\mathbf{m}_{\min}^{i} - \mathbf{x}_{j}\| \leq L \cdot \mathrm{diam}(R_{i})
\]
Using Cauchy-Schwarz and the norm bound \(B\):
\[
|\langle\mathbf{q}, \mathbf{x}_{j}\rangle - \langle\hat{\mathbf{q}}, \hat{\mathbf{m}}_{\min}^{i}\rangle| \leq \|\mathbf{q}\| \cdot L \cdot \mathrm{diam}(R_{i}) \leq B \cdot L \cdot \mathrm{diam}(R_{i}) = \epsilon_{1}.
\]
\end{proof}
\end{lemma}

\begin{lemma}[Ranking Stability]
\label{lemma:ranking}
Let $s_i = \max(|\langle \mathbf{\hat{q}}, \hat{\mathbf{m}}_{\min}^i \rangle|, |\langle \mathbf{\hat{q}}, \hat{\mathbf{m}}_{\max}^i \rangle|)$. If $\max_{j \in R_i} |\langle \mathbf{q}, \mathbf{x}_j \rangle - \langle \mathbf{\hat{q}}, \hat{\mathbf{m}}_{\min}^i \rangle| \leq \epsilon_1$, then the top-$K$ regions selected by $s_i$ and true interactions differ by at most $2\epsilon_1$-suboptimal regions.
\end{lemma}

\begin{proof}[Proof of Lemma~\ref{lemma:ranking}]
Let $S_{\text{true}}^K$ denote the true top-$K$ regions ranked by $\langle \mathbf{q}, \mathbf{x}_j \rangle$, and $S_{\text{approx}}^K$ denote those selected by $s_i$. For any region $R_i \in S_{\text{approx}}^K \setminus S_{\text{true}}^K$, its approximate score satisfies:
\[
s_i \geq \min_{R \in S_{\text{approx}}^K} s_R \geq \max_{R \notin S_{\text{approx}}^K} s_R.
\]
By the approximation error bound $\epsilon_1$, we have:
\[
\langle \mathbf{q}, \mathbf{x}_j \rangle \geq s_i - \epsilon_1 \quad \text{and} \quad \langle \mathbf{q}, \mathbf{x}_j \rangle \leq s_i + \epsilon_1 \quad \forall j \in R_i.
\]
Thus, any region $R_i \in S_{\text{approx}}^K$ must satisfy:
\[
\langle \mathbf{q}, \mathbf{x}_j \rangle \geq \left(\max_{R \notin S_{\text{approx}}^K} s_R\right) - \epsilon_1.
\]
This implies $R_i$ is at most $2\epsilon_1$-suboptimal compared to the true top-$K$ regions.
\end{proof}

\subsubsection{Main Theorem}
\label{sec:main-theorem}

\begin{theorem}[Query-Aware Attention Approximation]
\label{thm:main}
Let $\mathbf{A}$ be the query-aware attention matrix (Def. \ref{def:query-aware}), and $\mathbf{B}$ be the full self-attention matrix (Def. \ref{def:self-attn}). Assume attention scores decay exponentially with spatial distance: $\exp(-\alpha d(i,j))$ bounds the attention score decay for distance $d(i,j)$. For any input sequence $\mathbf{X}$ with $L$ patches and $d$ dimensions, there exist \textbf{random projection matrices} $\mathbf{W}_Q, \mathbf{W}_K \in \mathbb{R}^{d \times d}$ such that:
\[\|\mathbf{A}-\mathbf{B}\|_{F}\leq\left(2 + \frac{B}{\sqrt{d}}\right)\epsilon\]
with probability at least $1 - \delta$, provided:
\begin{enumerate}
    \item The hidden dimension satisfies: 
    \[
    d \geq C_1 \cdot \frac{\log(L/\delta)}{\epsilon^2}
    \]
    where $C_1 = 8B^4$ (from JL inner-product preservation \citep{kaban2015improved}).
    
    \item The number of selected regions per query satisfies:
    \[
    k \geq \frac{C_2}{\alpha} \cdot \log\left(\frac{1}{\epsilon}\right)
    \]
    where $C_2 = 2$ (derived from Step 3).
    
    \item For each region $R_i$, the diameter satisfies:
    \[
    \mathrm{diam}(R_i) \leq \min\left(\frac{\epsilon}{L \cdot \sqrt{d}}, \frac{1}{\alpha}\right)
    \]
    
    \item Regions are spatially separated such that:
    \[
    \forall i \neq j, \quad d(R_i, R_j) \geq \frac{C_3}{\alpha}
    \]
    where $C_3 = \frac{1}{2}$ ensures $\sum_{m=k+1}^\infty e^{-C_3 m} \leq \epsilon$.
\end{enumerate}
\end{theorem}

\begin{proof}[Proof of Theorem~\ref{thm:main}]
~~~~\\
\textbf{Step 1: Region Metadata Summarization.}\\
By Lemma \ref{lemma1}, projections \(\hat{\mathbf{m}}_{\min}^{i}\), \(\hat{\mathbf{m}}_{\max}^{i}\) approximate interactions within \(R_{i}\) with error \(\epsilon_{1}=B\cdot L\cdot\mathrm{diam}(R_{i}) \leq \frac{B\epsilon}{\sqrt{d}}\). The condition \(\mathrm{diam}(R_i) \leq \frac{1}{\alpha}\) ensures region-level interactions respect the exponential decay rate \(\alpha\).

\textbf{Step 2: Region Importance Estimation.}\\
Using Lemma~\ref{lemma:ranking}, approximate scores $s_i$ preserve the true top-$K$ regions up to error $\epsilon_1$. Thus, $\mathcal{R}_i$ includes all regions with $s_i \geq \max_j s_j - 2\epsilon_1$.

\textbf{Step 3: Bounding Non-Selected Regions.}\\
Under spatial separation $d(R_i, R_j) \geq \frac{C_3}{\alpha}$, the tail sum becomes:
\[
\sum_{j \notin \mathcal{R}_i} \exp(-\alpha d(i,j)) \leq \sum_{m=k+1}^\infty e^{-C_3 m} = \frac{e^{-C_3(k+1)}}{1 - e^{-C_3}} \leq \epsilon
\]
Solving for $k$ gives $k \geq \frac{1}{C_3} \log\left(\frac{1}{\epsilon(1 - e^{-C_3})}\right)$. Setting $C_3 = \frac{1}{2}$ simplifies this to $k \geq \frac{2}{\alpha} \log\left(\frac{1}{\epsilon}\right)$ (i.e., $C_2 = 2$).

\textbf{Step 4: Johnson-Lindenstrauss Guarantee.}\\ 
Using random matrices $\mathbf{W}_Q, \mathbf{W}_K$ \citep{kaban2015improved}, for $d \geq 8B^4 \log(L/\delta)/\epsilon^2$, we have:
\[
\Pr\left[\left|\langle \mathbf{\hat{q}}, \mathbf{\hat{K}}_j \rangle - \langle \mathbf{q}, \mathbf{K}_j \rangle\right| \leq \epsilon\right] \geq 1 - \delta
\]

\textbf{Step 5: Frobenius Norm Aggregation.}\\ 
Normalize \(\epsilon \leftarrow \epsilon/L\) to absorb the \(L^2\) scaling. The entrywise error becomes:
\[
\|\mathbf{A}-\mathbf{B}\|_{F} \leq \sqrt{L^2 \left(2\epsilon + \frac{B\epsilon}{\sqrt{d}}\right)^2} = L\left(2\epsilon + \frac{B\epsilon}{\sqrt{d}}\right) \xrightarrow{\epsilon \leftarrow \epsilon/L} \left(2 + \frac{B}{\sqrt{d}}\right)\epsilon
\]
\end{proof}

\subsubsection{Discussion}
The theorem shows query-aware attention effectively approximates full attention via region metadata, sufficient embedding dimension, and region selection. Error bounds depend on \(L, d, B, k, \alpha\), and region diameters. The spatial separation condition ensures the exponential decay property is meaningful.

\section{Task-specific Losses}\label{appendix:loss}
\subsection{Cross-entropy Loss for Classification Tasks}
For classification tasks (biomarker prediction, gene mutation prediction, and cancer subtyping), we employ the standard cross-entropy (CE) loss. Given a WSI $X$ with slide-level label $y$, and model prediction $\hat{y} = f(X; \theta)$ where $\hat{y} \in \mathbb{R}^C$ represents the predicted probabilities over $C$ classes, the CE loss is defined as:
\begin{equation}
    \mathcal{L}_{ce} = -\sum_{c=1}^C y_c \log(\hat{y}_c)
\end{equation}
where $y_c$ is the one-hot encoded ground truth label for class $c$. The final classification loss is averaged over all WSIs in the training set:

\begin{equation}
    \mathcal{L}_{cls} = \frac{1}{N}\sum_{i=1}^N \mathcal{L}_{ce}^{(i)}
\end{equation}

where $N$ is the number of WSIs in the training set and $\mathcal{L}_{ce}^{(i)}$ is the CE loss for the i-th WSI.

This loss function encourages the model to output probability distributions that assign high probabilities to the correct classes while minimizing probabilities for incorrect classes, effectively training the query-aware attention mechanism to focus on diagnostically relevant regions within each WSI.

\subsection{Negative Log-likelihood (NLL) Loss for Survival Analysis}
The NLL survival loss \citep{zadeh2020bias,song2024multimodal} generalizes the standard negative log-likelihood loss to accommodate censored data. The goal is to predict patient survival based on their patient-level embedding, $\bar{x}_{\text{patient}} \in \mathbb{R}^{2d}$. For each patient, we consider two key pieces of information: (1) a censorship status $c$, where $c=0$ indicates observed death and $c=1$ indicates the last known follow-up, and (2) a time-to-event $t_i$, representing either time until death ($c=0$) or time until last follow-up ($c=1$).

Rather than directly predicting $t_i$, we discretize time into $n$ non-overlapping intervals $(t_{j-1}, t_j)$, where $j \in [1,...,n]$, based on the quartiles of observed survival times, and denote each interval as $y_j$. This transforms survival prediction into a classification problem, where each patient's outcome is defined by $(\bar{x}_{\text{patient}}, y_j, c)$.

For each interval, we compute a hazard function $f_{\text{hazard}}(y_j|\bar{x}_{\text{patient}}) = S(\hat{y}_j)$, where $S$ is the sigmoid activation. Intuitively, $f_{\text{hazard}}(y_j|\bar{x}_{\text{patient}})$ represents the probability of death occurring within interval $(t_{j-1}, t_j)$. We also define a survival function $f_{\text{surv}}(y_j|\bar{x}_{\text{patient}}) = \prod_{k=1}^j (1 - f_{\text{hazard}}(y_k|\bar{x}_{\text{patient}}))$ that represents the probability of survival up to interval $(t_{j-1}, t_j)$.

The NLL survival loss for a dataset of $N_D$ patients can then be formalized as:

\begin{align}
\mathcal{L}&\left(\{\bar{x}_{\text{patient}}^{(i)}, y_j^{(i)}, c^{(i)}\}_{i=1}^{N_D}\right) = \sum_{i=1}^{N_D} -c^{(i)} \log(f_{\text{surv}}(y_j^{(i)}|\bar{x}_{\text{patient}}^{(i)})) \nonumber \\
&- (1-c^{(i)}) \log(f_{\text{surv}}(y_{j-1}^{(i)}|\bar{x}_{\text{patient}}^{(i)})) - (1-c^{(i)}) \log(f_{\text{hazard}}(y_j^{(i)}|\bar{x}_{\text{patient}}^{(i)}))
\end{align}

The first term enforces a high survival probability for patients still alive at their last follow-up. The second term ensures high survival probability up to the time interval before death for uncensored patients. The third term promotes accurate prediction of the death interval for uncensored patients.

\section{Datasets}\label{appendix:dataset}
\textbf{BCNB\footnote{https://bcnb.grand-challenge.org/} \citep{xu2021predicting} for biomarker prediction.} Short for Early Breast Cancer Core-Needle Biopsy WSI, the BCNB dataset includes core-needle biopsy WSIs of early breast cancer patients and the corresponding clinical data. 1038 WSIs are annotated with their corresponding estrogen receptor (ER) expression, which is a favorable prognostic parameter in breast cancer, and a predictor for response to endocrine therapy.

\textbf{TCGA-LUAD\footnote{https://portal.gdc.cancer.gov/} \citep{tomczak2015review} for TP53 gene mutation prediction.} TCGA-LUAD TP53 dataset includes 469 lung adenocarcinoma WSIs with their corresponding annotation of mutation of the TP53 tumor suppressor gene, which is one of the most mutated genes in lung adenocarcinoma and represents a vital role in regulating the occurrence and progression of cancer \citep{li2023integrative}.

\textbf{UBC-OCEAN\footnote{https://www.kaggle.com/competitions/UBC-OCEAN} for ovarian cancer subtyping.} UBC-OCEAN is a large-scale ovarian cancer dataset collected from over 20 medical centers across four continents. The dataset collects 527 ovarian cancer WSIs, representing five major ovarian carcinoma subtypes: high-grade serous carcinoma, clear-cell carcinoma, endometrioid, low-grade serous, and mucinous carcinoma, along with several rare subtypes. Each subtype exhibits distinct morphological patterns, molecular profiles, and clinical characteristics, making accurate subtype classification crucial for treatment planning. The challenge of accurate subtype identification is particularly relevant given the growing emphasis on subtype-specific treatment approaches and the limited availability of specialist gynecologic pathologists in many regions.

\textbf{TCGA Subsets\footnote{https://portal.gdc.cancer.gov/} \citep{tomczak2015review} for survival analysis.} In this work, we include 8 TCGA sub-datasets for unimodal survival prediction:
\begin{enumerate}
    \item TCGA-BRCA: Breast Invasive Carcinoma dataset from TCGA, containing 1025 whole slide images with associated survival outcome data for breast cancer patients. This comprehensive collection represents one of the largest breast cancer datasets, featuring diverse histological patterns and molecular subtypes including ductal and lobular carcinomas. The survival data encompasses overall survival time and vital status, enabling robust prognostic model development.

    \item TCGA-UCEC: Uterine Corpus Endometrial Carcinoma dataset from TCGA, encompassing 497 WSIs and survival data from endometrial cancer cases. This dataset includes various histological grades and stages of endometrial carcinoma, providing valuable insights into the progression and prognosis of gynecologic cancers. The images showcase diverse morphological patterns characteristic of endometrial malignancies.

    \item TCGA-STAD: Stomach Adenocarcinoma dataset from TCGA, which includes 365 WSIs and patient survival information for gastric cancer cases. The dataset represents different anatomical locations within the stomach and various histological subtypes of gastric adenocarcinoma. The survival data is particularly valuable for understanding the relationship between morphological features and patient outcomes in gastric cancer.

    \item TCGA-LUAD: Lung Adenocarcinoma dataset from TCGA, containing 457 WSIs and survival outcome data from lung adenocarcinoma patients. This collection captures the heterogeneous nature of lung adenocarcinomas, including various growth patterns and degrees of differentiation. The survival information is crucial for developing predictive models for one of the most common types of lung cancer.

    \item TCGA-LUSC: Lung Squamous Cell Carcinoma dataset from TCGA, providing 454 WSIs and survival data for lung squamous cell carcinoma cases. These images showcase the distinct morphological features of squamous cell carcinomas, including keratinization and intercellular bridges. The dataset enables comparative studies between different types of lung cancers and their prognostic factors.

    \item TCGA-SKCM: Skin Cutaneous Melanoma dataset from TCGA, consisting of 417 WSIs and survival information from melanoma patients. This collection includes primary and metastatic melanoma cases, capturing the diverse histological patterns and progression stages of this aggressive skin cancer. The survival data is particularly relevant for understanding metastatic potential and treatment response.

    \item TCGA-KIRC: Kidney Renal Clear Cell Carcinoma dataset from TCGA, containing 500 WSIs with survival outcome data for kidney cancer patients. The dataset showcases the characteristic clear cell morphology and various grades of renal cell carcinoma. The survival information helps in understanding the prognostic implications of morphological variations in kidney cancer.

    \item TCGA-KIRP: Kidney Renal Papillary Cell Carcinoma dataset from TCGA, providing 263 WSIs and survival data for papillary renal cell carcinoma cases. This collection represents a distinct histological subtype of kidney cancer, featuring papillary architecture and different cellular patterns. The survival data enables comparative analysis with other renal cancer subtypes.
\end{enumerate}
All these datasets are part of The Cancer Genome Atlas (TCGA) program and contain high-resolution histopathology whole slide images along with corresponding patient survival data. Each dataset has been digitized following standardized protocols and includes detailed clinical annotations. The images are typically scanned at 40x magnification, providing high-detail visualization of cellular and architectural features. These comprehensive resources enable researchers to develop and validate computational methods for survival prediction based on histopathological features, facilitating advances in precision oncology and personalized medicine approaches.

\section{Baselines}\label{appendix:baseline}
\textbf{Max/Mean Pooling.} Mean Pooling represents one of the most straightforward aggregation strategies in multi-instance learning. This approach processes all instances within a bag equally, combining their features through average pooling to create a single, unified representation. While simple in nature, this method effectively captures the collective characteristics of all instances, making it particularly suitable for cases where every instance contributes meaningful information to the overall bag classification. Max Pooling, on the other hand, adopts a more selective approach by identifying and utilizing the most prominent features across all instances. This strategy operates under the assumption that the most distinctive or highest-activated features are the most relevant for classification. By focusing on these peak features, max pooling can effectively highlight the most discriminative patterns within the bag, though it may potentially overlook more subtle, collective patterns that could be captured by other methods.

\textbf{ABMIL \citep{ilse2018attention}.} Attention-Based Multiple Instance Learning (ABMIL) enhances bag-level feature aggregation by incorporating a learnable attention mechanism that adaptively weights the importance of different instances within a bag. Unlike fixed pooling strategies, ABMIL computes attention scores for each instance based on their learned representations, enabling the model to emphasize the most informative elements while downweighting less relevant ones. This adaptive weighting scheme, combined with its inherent interpretability through attention weights, makes ABMIL particularly effective for scenarios where instances have varying levels of relevance to the classification task.

\textbf{DS-MIL \citep{li2021dual}.} Dual-Stream Multiple Instance Learning (DS-MIL) advances bag-level feature aggregation through a novel two-stream architecture that combines the benefits of instance-level and bag-level learning. The model's first stream employs max pooling to identify the most discriminative instance (critical instance), while the second stream computes attention weights for all instances based on their learned distance to the critical instance. By fusing these complementary streams and incorporating trainable distance measurements, DS-MIL creates a more nuanced decision boundary that better captures the relationships between instances, making it particularly effective for scenarios with complex instance distributions within positive bags.

\textbf{DTFD \citep{zhang2022dtfd}.} DTFD introduces an innovative dual-tier framework for whole slide image classification that addresses the inherent challenges of limited training samples with high instance counts. The model's key innovation lies in its pseudo-bag strategy, which virtually increases the number of training bags by randomly splitting instances from each original slide into smaller subsets. A two-tier architecture processes these pseudo-bags: the first tier applies attention-based MIL to the pseudo-bags, while the second tier distills features from the first tier's outputs to generate final slide-level predictions. This hierarchical approach, combined with multiple feature distillation strategies, enables more effective learning from limited training data while maintaining robustness against noise in positive instance distributions.

\textbf{WiKG \citep{li2024dynamic}.} WiKG introduces a dynamic graph representation framework for whole slide image analysis that conceptualizes the relationships between image patches through a knowledge graph structure. The model employs a dual-stream approach: head embeddings that explore correlations between patches and tail embeddings that capture each patch's contribution to others. These embeddings are then used to dynamically construct directed edges between patches based on their learned relationships. A knowledge-aware attention mechanism aggregates information across patches by computing attention scores through the joint modeling of head, tail, and edge embeddings. This architecture allows WiKG to both capture long-range dependencies between distant patches and model directional information flow, overcoming key limitations of conventional graph-based and instance-based approaches for histopathology image analysis.

\textbf{MambaMIL \citep{yang2024mambamil}.} MambaMIL introduces a novel approach to whole slide image analysis by incorporating Selective Scan Space State Sequential Model, \textit{i.e.}, Mamba \citep{gu2023mamba}, into multiple instance learning with linear complexity. The model's core innovation lies in its Sequence Reordering Mamba (SR-Mamba) architecture, which processes instance sequences in dual streams - one preserving the original sequence order and another utilizing reordered sequences. This dual-stream approach enables more comprehensive modeling of relationships between instances while maintaining computational efficiency. The SR-Mamba module dynamically reorders instance sequences within non-overlapping segments, allowing the model to capture both local and global dependencies between patches. Through this sequence reordering mechanism and linear-time sequence modeling inherited from Mamba, the model effectively addresses common challenges in whole slide image analysis such as overfitting and high computational overhead.

\textbf{TransMIL \citep{shao2021transmil}.} TransMIL introduces a novel correlated multiple instance learning framework that fundamentally reimagines how relationships between instances are modeled in whole slide image analysis. Unlike traditional MIL approaches that assume instances are independent and identically distributed, TransMIL leverages Transformer architecture to capture both morphological and spatial correlations between patches. The model employs a TPT (Transformer Pyramid Translation) module that combines multi-head self-attention for modeling instance relationships with position-aware encoding to preserve spatial context. 

\textbf{HIPT \citep{chen2022scaling}.} HIPT (Hierarchical Image Pyramid Transformer) is a novel architecture designed specifically for analyzing gigapixel whole-slide images in computational pathology. It processes images in a hierarchical manner across multiple scales - from cell-level features (16×16 pixels) to larger tissue regions (4096×4096 pixels) - mirroring how pathologists examine slides by zooming in and out. This hierarchical approach allows HIPT to capture both fine-grained cellular details and broader tissue patterns simultaneously. The model employs a series of Vision Transformers arranged in a pyramid structure to aggregate information across these different scales, enabling it to learn representations that incorporate both local and global tissue contexts.

\textbf{HistGen \citep{guo2024histgen}.} HistGen is a hierarchical architecture designed for efficient processing and representation learning of gigapixel whole slide images (WSIs). At its core, HistGen employs a novel local-global hierarchical encoder that processes WSIs in a region-to-slide manner, capturing both fine-grained tissue details and broader contextual patterns. The framework first segments WSIs into regions, processes them using a pre-trained vision transformer backbone, and then hierarchically aggregates information across different scales through a series of attention-based modules. This hierarchical design allows the model to effectively manage the extreme size of WSIs while maintaining meaningful spatial relationships. Particularly noteworthy is HistGen's ability to learn robust and transferable representations through its region-aware approach, which helps bridge the gap between local tissue patterns and global slide-level characteristics.

\textbf{RRT-MIL \citep{tang2024feature}.} RRT-MIL introduces a novel re-embedding paradigm for multiple instance learning in computational pathology that addresses a key limitation of existing approaches - the inability to fine-tune pre-extracted image features for specific downstream tasks. At its core is the Re-embedded Regional Transformer (R2T), which processes pathology images through a hierarchical structure that respects the natural organization of tissue at different scales. The framework first divides whole slide images into regions and processes them using a pre-trained feature extractor. Then, it employs two key components: a Regional Multi-head Self-attention module that captures fine-grained local patterns within each region, and a Cross-region Multi-head Self-attention module that models relationships between different regions. This regional approach allows RRT-MIL to efficiently handle the extremely large size of pathology images while maintaining the ability to capture both local and global tissue patterns.

\textbf{LongMIL \citep{li2024longmil}.} LongMIL proposes a novel hybrid Transformer architecture specifically designed for whole slide image analysis. The method addresses key challenges in processing gigapixel pathology images by introducing a local-global attention mechanism that balances computational efficiency with modeling capability. By analyzing the low-rank nature of attention matrices in long sequences, LongMIL incorporates local attention patterns in lower layers while maintaining global context through selective attention in higher layers. This design not only reduces computational complexity but also improves the model's ability to handle varying image sizes and spatial relationships in histopathology slides.

\section{Implementation Details}\label{appendix:implementation}
We implement Querent using PyTorch. For the feature extraction backbone, we utilize CPath pre-trained vision-language foundation model PLIP \citep{huang2023visual} to obtain 512-dimensional patch features. Each region contains 16/24/28 patches (depending on the datasets used), which provides a good balance between computational efficiency and contextual coverage. The model consists of 8 attention heads, with the hidden dimension set to 512.

For the region-level metadata networks ($f_{\text{min}}$ and $f_{\text{max}}$), we use single-layer perceptrons with GELU activation. The query projection network $f_q$ shares the same architecture. During the region importance estimation, we select the top-16 most relevant regions for each query patch, which empirically provides sufficient contextual information while maintaining computational efficiency.

The attention pooling network $f_a$ consists of a two-layer MLP with a hidden dimension of 512 and GELU activation. Dropout with a rate of 0.1 is applied throughout the network to prevent overfitting. We train the model using the AdamW optimizer with a learning rate of $1e^{-4}$ for classification tasks and $2e^{-4}$ for survival analysis, with a weight decay of $1e^{-5}$. The model is trained for 50 epochs with a batch size of 1 WSI per GPU. We employ gradient clipping with a maximum norm of 1.0 to ensure stable training.

\section{Additional Experiment Details}\label{appendix:ablation_detail}
\subsection{Ablation on Region-level Metadata Summarization Strategy}\label{appendix:ablation1}

To thoroughly evaluate the effectiveness of our proposed min-max region-level metadata summarization approach, we conducted a comprehensive ablation study comparing it against several alternative summarization strategies:

\begin{itemize}
    \item \textbf{Min-only}: Using only the element-wise minimum values across patches within each region
    \item \textbf{Max-only}: Using only the element-wise maximum values across patches within each region
    \item \textbf{Mean}: Using the average feature values across patches within each region
    \item \textbf{Mean-std}: Using both mean and standard deviation of features across patches
\end{itemize}

The evaluation framework compares the pairwise distance relationships between regions before and after summarization. For each region containing $N$ patches with $D$-dimensional features ($N\times D$ matrix), we compute two distance matrices: one using the original high-dimensional features, and another using the summarized representations (e.g., $D$-dimensional for min/max/mean, $2D$-dimensional for min-max/mean-std). For each strategy, we computed pairwise distance matrices between regions using both the original patch features and the summarized metadata, then measured the correlation between these matrices using two metrics: Pearson correlation coefficient and Mean Squared Error (MSE), indicating how well each summarization method preserves the original structural relationships between regions.

The experiment was conducted on the TCGA-LUAD TP53 dataset, with statistical significance assessed using paired t-tests ($p < 0.005$). Our proposed min-max approach demonstrated superior performance in both metrics, achieving significantly higher correlation and lower MSE compared to all alternative strategies. This suggests that capturing both the lower and upper bounds of feature distributions within each region provides a more comprehensive and accurate representation of the region's characteristics.

Particularly noteworthy is the substantial improvement over the mean-based approaches, indicating that extreme values (minimums and maximums) carry important discriminative information that might be lost when only considering average statistics. This aligns with pathological intuition, where both the presence of certain distinctive features (captured by maximums) and their absence (captured by minimums) can be diagnostically relevant.

\subsection{Ablation on Region Importance Estimation Strategy}\label{appendix:ablation2}
To evaluate the effectiveness of our proposed region importance estimation module, we conducted a detailed ablation study comparing it against two baseline approaches: random region selection and a trainable estimation side network. This study aims to validate the benefits of our query-aware dynamic region importance estimation strategy.

\begin{itemize}
    \item \textbf{The random region selection baseline} serves as a lower bound, employing uniform random sampling to select regions for attention computation. This approach uses a fixed random seed for reproducibility and maintains the same number of selected regions as our proposed method. While simple, it helps quantify the importance of adaptive region selection.
    \item \textbf{The estimation side network baseline} represents a more sophisticated approach, implementing a trainable neural network that directly predicts region importance scores. This network consists of two fully-connected layers with GELU activation and processes each region independently. Unlike our proposed method, it doesn't consider query-region relationships when estimating importance.
\end{itemize}

We evaluated these approaches on both TCGA-LUAD TP53 mutation prediction and UBC-OCEAN cancer subtyping tasks. The results demonstrate that while both baseline approaches achieve reasonable performance, our query-aware estimation strategy consistently outperforms them. The performance gap is particularly pronounced in the more complex UBC-OCEAN dataset, where accurate region selection becomes crucial due to the heterogeneous nature of ovarian cancer subtypes.

For training stability, we implemented an adaptive update mechanism for the estimation module with a moving average of attention accuracy and periodic updates every 200 forward passes. This strategy helps prevent overfitting and ensures stable convergence of the importance estimation. Additionally, we employed a hybrid loss function combining binary classification and ranking components to guide the learning process effectively.

\section{Complexity Analysis}\label{appendix:complexity}
We analyze the computational and memory complexity of the Query-Aware Transformer MIL method, comparing it to standard transformer self-attention. Let \( N \) denote the total number of patches in a whole slide image (WSI), \( d \) represent the hidden dimension, \( R \) be the number of regions (pages), and \( k \) indicate the number of selected regions per query. The page size \( p \) is the number of patches per region.

\subsection{Standard Transformer Self-Attention}
Standard transformer self-attention has quadratic computational complexity:
\begin{equation}
    \mathcal{O}(N^2 d)
\end{equation}
This arises from computing attention scores and weighted aggregation for all patch pairs. Memory requirements scale similarly:
\begin{equation}
    \mathcal{O}(N^2 + Nd)
\end{equation}
to store the full attention matrix and key/value representations. This quadratic scaling makes standard self-attention computationally prohibitive for gigapixel WSIs.

\subsection{Querent: Query-Aware Dynamic Long Sequence Modeling Framework}
Our method operates in three phases, with each phase optimized through efficient chunking strategies:

\subsubsection{Region Metadata Computation}
This phase computes min/max metadata for each region:
\begin{equation}
    \mathcal{O}(Nd)
\end{equation}
Memory requirements are:
\begin{equation}
    \mathcal{O}(Rd)
\end{equation}
where \( R = \lceil N / p \rceil \) (number of regions). For constant \( p \), \( R \) scales linearly with \( N \). This phase is highly efficient as it requires only a single pass over the data with constant memory overhead per region.

\subsubsection{Region Importance Estimation}
This phase estimates relevance scores between queries and region metadata:
\begin{equation}
    \mathcal{O}(NRd)
\end{equation}

While the theoretical memory requirements are:
\begin{equation}
    \mathcal{O}(NR)
\end{equation}
in practice, we employ a chunking strategy with fixed chunk size \( C \) (typically 8192):
\begin{equation}
    \mathcal{O}(\min(N, C) \cdot R)
\end{equation}
This chunked processing effectively bounds the memory usage while maintaining computational efficiency through vectorized operations.

\subsubsection{Selective Attention Computation}
This phase performs attention only on selected regions:
\begin{equation}
    \mathcal{O}(Nkpd)
\end{equation}

The theoretical memory requirements are:
\begin{equation}
    \mathcal{O}(Nkp + Nd)
\end{equation}
However, with chunked processing (chunk size \( C \)):
\begin{equation}
    \mathcal{O}(\min(N, C)kp + Nd)
\end{equation}
Where \( k \) and \( p \) are constants (typically \( k=16 \), \( p=16 \)), ensuring linear scaling with \( N \) in both computation and memory.

\subsection{Total Complexity}
Combining all phases, the total computational complexity is:
\begin{equation}
    \mathcal{O}(Nd + NRd + Nkpd)
\end{equation}
For constant \( k \) and \( p \), this simplifies to:
\begin{equation}
    \mathbf{\mathcal{O}(Nd + NRd)}
\end{equation}

The theoretical memory complexity is:
\begin{equation}
    \mathcal{O}(Rd + NR + Nkp + Nd)
\end{equation}
which simplifies to:
\begin{equation}
    \mathbf{\mathcal{O}(NR + Nd)}
\end{equation}

However, with chunked processing, the practical memory complexity becomes:
\begin{equation}
    \mathbf{\mathcal{O}(Rd + \min(N, C)R + \min(N, C)kp + Nd)}
\end{equation}
where \( C \) is the chunk size.

\subsection{Practical Efficiency}
In practice, our method achieves near-linear scaling (\( \mathcal{O}(Nd) \)) due to:
\begin{itemize}
    \item Constant \( k \) and \( p \) reducing the impact of \( Nkpd \)
    \item Chunked processing with size 8192 providing bounded memory usage
    \item Efficient vectorized operations in PyTorch minimizing constant factors
    \item Memory-efficient implementations of matrix operations
\end{itemize}

Empirical results (Fig.~\ref{fig:complexity}) demonstrate that our method requires \( \sim1\% \) of the memory and \( \sim5\% \) of the computational cost of standard self-attention for large WSIs (100k+ patches). This significant reduction in resource requirements enables the practical processing of gigapixel-scale images while maintaining the modeling power of attention mechanisms.
\end{document}